# Multi-Fault Diagnosis Of Industrial Rotating Machines Using Data-Driven Approach: A Review Of Two Decades Of Research

Shreyas Gawde [1] · Shruti Patil [2*] · Satish Kumar [2] · Pooja Kamat[1] · Ketan Kotecha [2] · Ajith Abraham[3]

**Abstract**: Industry 4.0 is an era of smart manufacturing. Manufacturing is impossible without the use of machinery. Majority of these machines comprise rotating components and are called rotating machines. The engineers' top priority is to maintain these critical machines to reduce the unplanned shutdown and increase the useful life of machinery. Predictive maintenance (PdM) is the current trend of smart maintenance. The challenging task in PdM is to diagnose the type of fault. With Artificial Intelligence (AI) advancement, data-driven approach for predictive maintenance is taking a new flight towards smart manufacturing. Several researchers have published work related to fault diagnosis in rotating machines, mainly exploring a single type of fault. However, a consolidated review of literature that focuses more on "multi-fault diagnosis" of rotating machines is lacking. There is a need to systematically cover all the aspects right from sensor selection, data acquisition, feature extraction, multi-sensor data fusion to the systematic review of AI techniques employed in multi-fault diagnosis. In this regard, this paper attempts to achieve the same by implementing a systematic literature review on a Data-driven approach for multi-fault diagnosis of Industrial Rotating Machines using "Preferred Reporting Items for Systematic Reviews and Meta-Analysis" (PRISMA) method. The PRISMA method is a collection of guidelines for the composition and structure of systematic reviews and other meta-analyses. This paper identifies the foundational work done in the field and gives a comparative study of different aspects related to multi-fault diagnosis of industrial rotating machines. The paper also identifies the major challenges, research gap. It gives solutions using recent advancements in AI in implementing multi-fault diagnosis, giving a strong base for future research in this field.

[1] Symbiosis Institute of Technology (SIT), Symbiosis International (Deemed) University, Pune. Symbiosis (Deemed University), Pune, Maharashtra, 412115, India.

[2] Symbiosis Centre for Applied Artificial Intelligence (SCAAI), Symbiosis Institute of Technology, Pune. Symbiosis (Deemed University), Pune, Maharashtra, 412115, India.

[3] Machine Intelligence Research Labs, Auburn, WA 98071, USA

## 1. INTRODUCTION

The Industries are the basis of the nation's economy. With the recent advancement in technology, especially in Artificial Intelligence (AI), these industries have achieved new heights by transforming into smart factories, marking an era of the fourth industrial revolution. Smart factories employ smart manufacturing, and the basic building block of any manufacturing process industry is the machines. Most of the machines in the industries comprise rotating components and are called rotating machines [1]. More precisely, the rotating machines facilitate the transfer of energy to fluids and solids or vice versa. The rotating machine consists of the rotating part, which we call a rotor, and the static part called a stator [2]. In the process industry, a train of rotating machines is used to transport solids, liquids, and gases [3]. Process machinery comprises a group of sub-elements (of rotating machines) that combine to convert one form of energy until converted into the desired usable form of energy. There are different sub-elements, including the driver machines, the driven machines, the speed modifiers, the shaft, and the coupling. The driver machines take electrical, steam, or fluid energy and convert it into rotary power that can be used to drive a process machine. Electric motors, turbines, reciprocating engines (usage is very less) are examples of driver machines. Driven process machine transports a given process fluid or solid, at a given flow and pressure, to specific points in a process. Pumps, fans, compressors, conveyor belts, etc., are widely used driven machines. The speed of the driver output shaft may be increased or decreased by a speed modifier, depending on the requirement of the process machine being driven. Gearboxes, sheaves, and belts are examples of rotating machines that work as speed modifiers. However, Variable Frequency Drive (VFD) is electronic equipment used as a speed modifier. A shaft is a rotating machine element used to transmit energy from the driver to driven machinery. The shaft on the driver side is connected to the driven side using coupling. Fig. 1 shows the representation of rotating machines with Fig.1a [3] giving the flow of the energy is transfer in the process of rotating machines, and Fig. 1b gives an example of the train of rotating machines comprising an electric motor coupled to a pump.

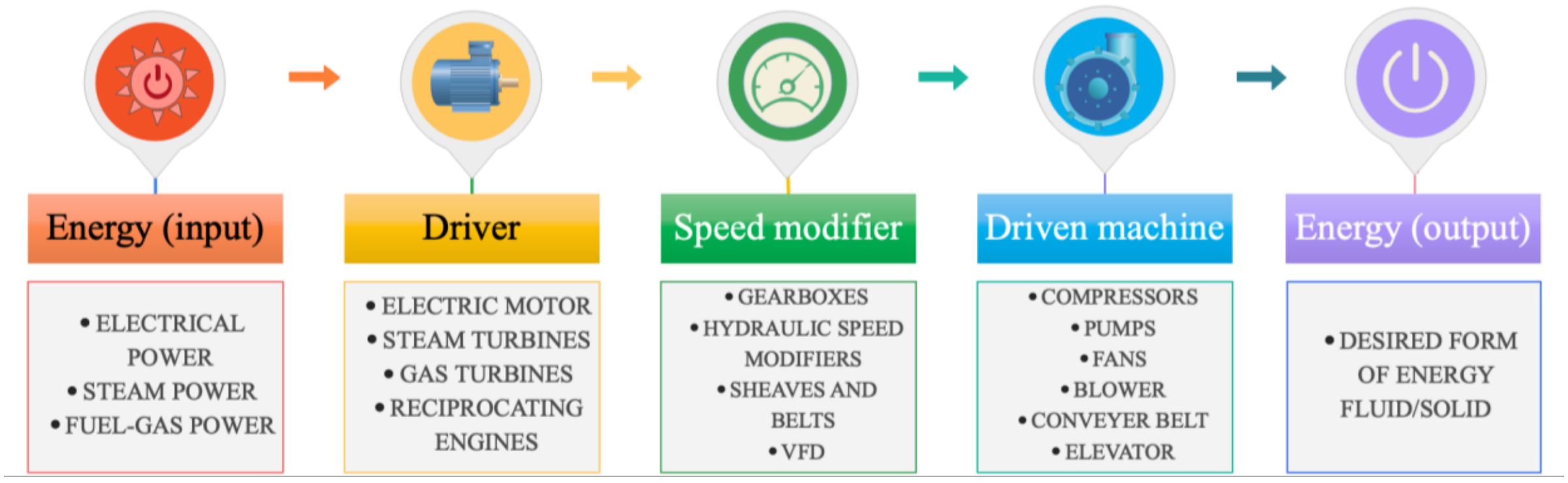

**a.** Flow Diagram of Process Rotating Machinery

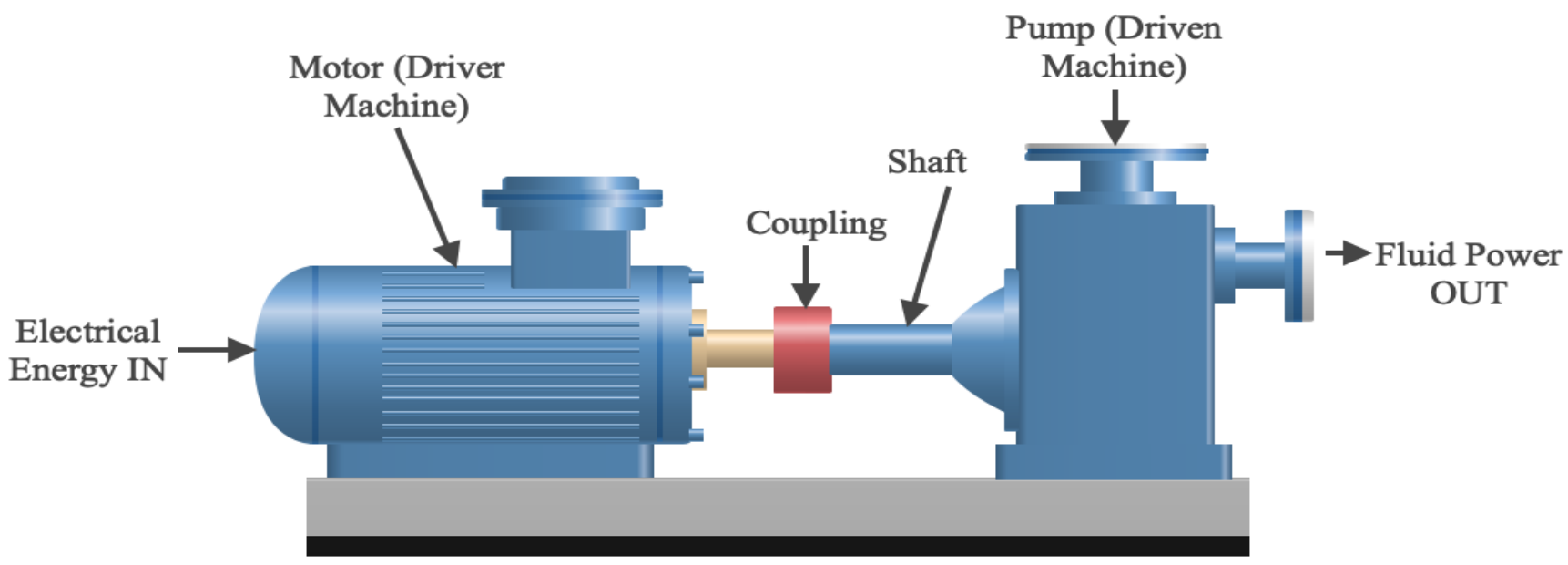

**b.** An Electric Motor coupled to a Centrifugal Pump.

**Figure 1.** Representation of Rotating Machines

From the above discussion, it is evident that any manufacturing process is incomplete without rotating machines. Therefore, it is essential to keep such machines in healthy operating conditions by deploying proper maintenance strategies [4]. Predictive maintenance is the current trend of smart maintenance, which most maintenance engineers follow. The challenging task in predictive maintenance is to diagnose the type of fault. With Artificial Intelligence (AI) advancement, a data-driven approach for predictive maintenance is taking a new flight towards smart manufacturing. The use of Big Data for multiple fault diagnoses will be the prime focus of the study.

## 1.1. Significance of the study

Rotating machines is the heart of any manufacturing process, and proper maintenance is the utmost priority of maintenance engineers. A proper maintenance strategy plays a vital role in the success of the manufacturing industry [5]. Figure 2 shows the P-F Curve [154], which depicts how equipment fails and how early identification of a failure allows time to plan and arrange the replacement or restoration of a failing item without causing a manufacturing slowdown. P denotes Potential failure (based on historical data), and F denotes Functional failure (actual failure) in a P-F Curve [152]. The main aim of any maintenance engineer is to maximize probability of failure using various maintenance strategies [153]. The figure also depicts the relation between the condition of the equipment and the cost to repair concerning the time taken to implement appropriate maintenance action. With the correct type of maintenance strategy, one can detect early failure mode.

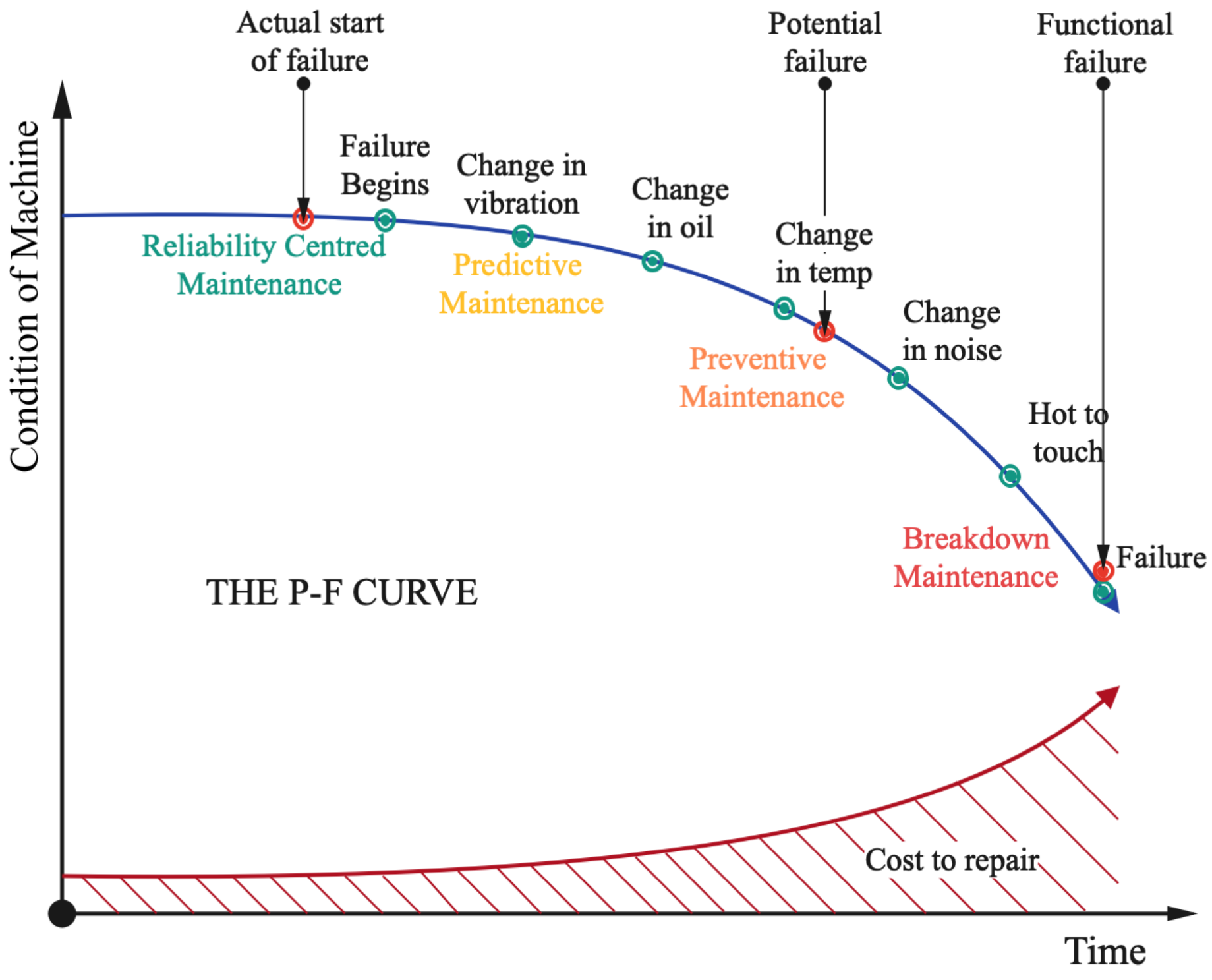

**Figure 2.** The PF Curve depicting the Significance of Early Fault Diagnosis.

It is very clear from figure 2 that it is crucial to have an early fault diagnosis to save the industrial economy. Digital transformation in Industry 4.0 has made it possible to collect a massive amount of data and effectively utilize it in fault diagnosis in Predictive Maintenance (PdM) [6]. It will reduce the unplanned downtime and increase the Remaining Useful Life (RUL) of the machinery [7]. With Artificial Intelligence (AI) advancement, a data-driven approach for predictive maintenance is taking a new flight towards smart manufacturing. Several researchers have published work related to fault diagnosis in rotating machines, mainly exploring a single type of fault. However, if we have to take complete advantage of Big Data, it is essential to not just focus on a single fault but to consider multiple faults that arise in the machinery [8]. The study is incomplete unless the AI models are generalized and detect maximum faults in real-time industrial environments with high prediction accuracy [9]. Also, most of the researchers have used single sensor data in fault diagnosis. A very few researchers have implemented multi-sensor data fusion, proving the drastic improvement in accuracy of diagnosis, taking into account the uncertainty of data [10]. Hence there is a need to consider this aspect of multi-sensor data fusion in future work. If a machine shows an unusual behavior or needs maintenance in the present scenario, a domain expert must diagnose the machinery's fault [11]. It shows that the research in this field still has a long future as a full-fledged implementation of the fault-diagnosis aspect using AI is yet to be achieved. A consolidated literature review is needed to focus on the "multi-fault diagnosis" aspect of rotating machines to give a strong foundation for future research. There is a need for a study that would systematically cover all the aspects right from sensor selection, data acquisition, feature extraction, multi-sensor data fusion to the systematic review of AI techniques employed in multiple fault diagnosis. This paper is also a small attempt to review the aspects mentioned above from the past research using PRISMA guidelines for systematic review.

## 1.2. Motivation

Artificial intelligence is an ocean of tremendous opportunities. The application of AI techniques to solve real-life problems is the new trend of future research. Being closely associated with the problems faced by maintenance engineers, finding a solution to the same utilizing AI techniques is the primary motivation behind the study. Fault detection is pretty simple by setting alarm limits, while a correct fault diagnosis is equally complex. With the advancements in Big data analysis using AI models, it is possible to tackle this problem to a great extent. However, most research focuses on the single type of fault, a preliminary study that needs further research. Also, the researchers have focussed on single sensor data, which again reduces the credibility of research as the health of machinery cannot be solely predicted based on a single sensor parameter. It is imperative to evaluate multiple parameters such as vibration, temperature, current, AE, etc., to get the complete condition of the machinery. To achieve this, multi-sensor data for fault diagnosis needs more exploration. The continuity in research lacks that would overcome the disadvantages of the previous work done. To explore the possible ways to solve the above problems, there is a need for an extensive literature survey. There is a need for study covering all aspects of multi-fault diagnosis, including big data acquisition, data processing, multi-sensor data fusion, and AI techniques that the researchers have already implemented. Such a study would give the gap in research and, in turn, motivate future research in the field. A very little encyclopedic research covered all the aforementioned aspects, which motivated the authors to explore more in this area. Fig.3 shows a trend in research in the past ten years extracted from the Scopus database in the field of multi-fault diagnosis in rotating machines. It is clear from the figure that the research is developing in this field, begging researchers' significant attention every year.

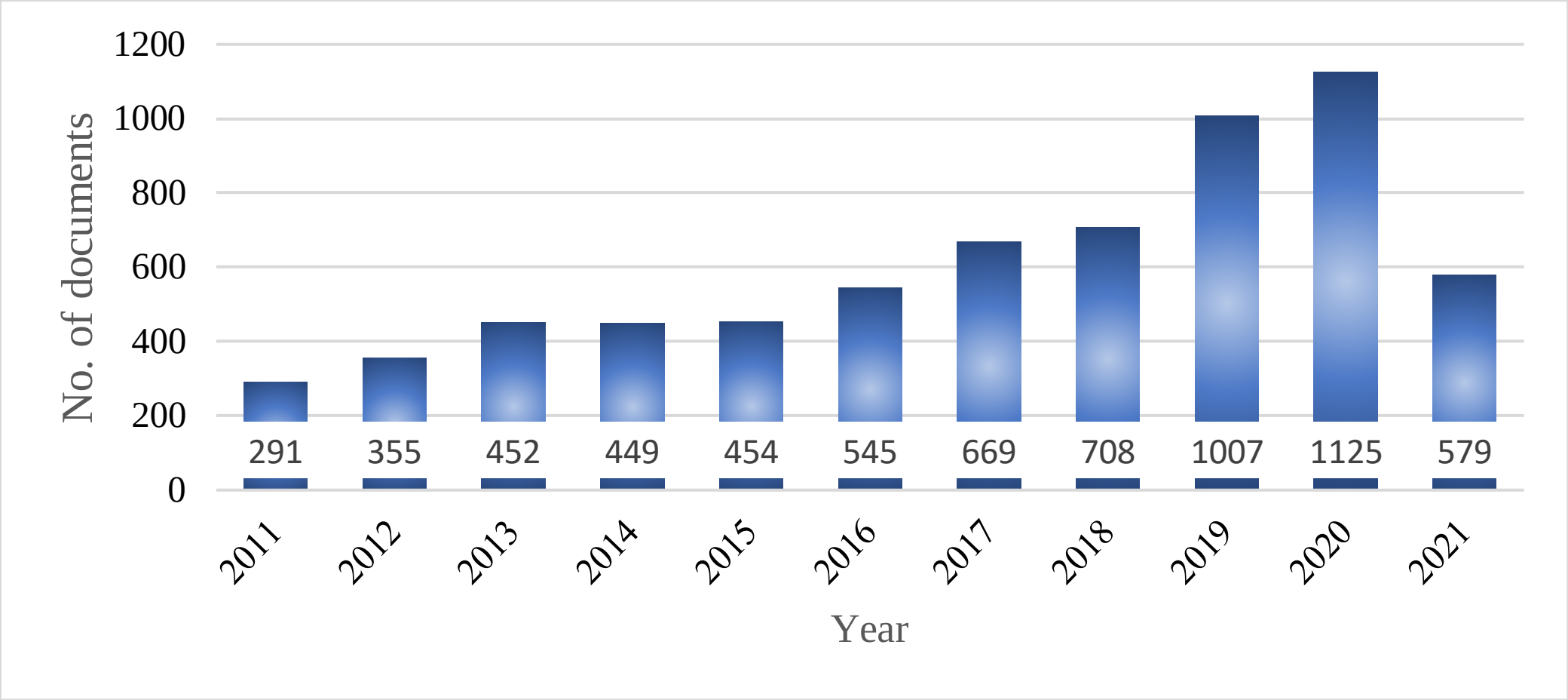

**Figure 3.** Research Trend in Fault Diagnosis of Rotating Machines over Past Ten Years (2011-2021)

## 1.3. Terms and Terminology

Following are the terms frequently used in fault-diagnosis of Industrial rotating machines.

- Industry 4.0: Industry 4.0 (the fourth industrial revolution) is the current automation trend, a fusion of cyber-physical systems, the Internet of things, and cloud computing [12].
- Predictive maintenance (PdM): In predictive maintenance, engineers try to predict the failures of machines based on certain conditions prevailing in the machinery.
- Big-Data Analysis: Big data analytics uses advanced analytic techniques against vast, diverse data sets that include structured, semi-structured, and unstructured data from different sources and in different sizes, from terabytes to zettabytes.
- Artificial Intelligence (AI): A subset of big data where we simulate human intelligence on machines.

- Data-driven approach: A data-driven approach gives decisions based on complex data analysis and interpretation rather than observation.

- Multi-sensor data fusion: Sensor fusion is the process of combining sensory data or data derived from disparate sources such that the resulting information has less uncertainty than would be possible when these sources used individually.

- Multi-fault diagnosis: Diagnosing multiple faults in the machinery based on data processing and analysis.

- Rotating machines: Rotating machines are generally used in the oil and gas and process industries to describe mechanical components that use kinetic energy to move fluids, gases, and other process materials.

- Maintenance strategy: A maintenance strategy defines the rules for the sequence of planned maintenance work.

## 1.4. Evolution of Maintenance Strategies:

Rotating machinery is a machine with a rotating component that transfers energy to a fluid, solid, or vice versa. In the first section of the introduction, we have discussed the different types of rotating machines, including the driver and the driven rotating machines. We have also seen the significance of these machines for the manufacturing process. Hence, maintenance engineers' most important task is to keep them in a healthy working condition. Let us understand the various maintenance philosophies that have evolved over the years.

First on the list is the Run to failure or breakdown maintenance philosophy. The machine runs in its prevailing condition without any check-up, and maintenance will be carried out only after breakdown or failure. It is also often referred to as reactive or corrective maintenance [1]. The temporary advantage is that it needs the least planning and hence no initial maintenance costs. On the other hand, it is expensive as the engineer waits until the end and acts based on the breakdown. Breakdown maintenance is an age-old technique but still followed in some machines, which are not so critical to the functioning of the plant. Demand for more complex machines due to increased rate of productivity made breakdown maintenance quite a lot expensive. At this point, another frequently used philosophy called preventive maintenance was introduced [1,13]. Few commonly used synonyms for preventive maintenance are scheduled, planned, and sometimes calendar-based maintenance. As the name again suggests, the engineers try to prevent failure by periodic or planned maintenance activities. It is not wrong from a machinery health perspective, but it becomes costly considering the resource (time and money) consumption. Hence preventive maintenance is also avoided as it results in increased maintenance costs though ensuring excellent and stable machinery health. So the engineers thought of changing this calendar-based maintenance to condition-based maintenance, which is called Condition Monitoring. Condition monitoring (CM) is the technique of continuously monitoring a machine's condition parameter (vibration, temperature, etc.) to detect a substantial change that might indicate a growing defect.

In the present era of Industry 4.0, engineers came up with another logical approach: the philosophy of predictive maintenance [14]. Predictive Maintenance is based on Condition Monitoring, abnormality detection, and AI algorithms. It integrates predictive models that can estimate the remaining machine runtime left or diagnose the type of fault in machinery according to detected abnormalities. This approach uses a wide range of tools, such as statistical analyses and Machine Learning to predict the state of the equipment. The evolution of Maintenance strategies is as shown in fig. 4 [7].

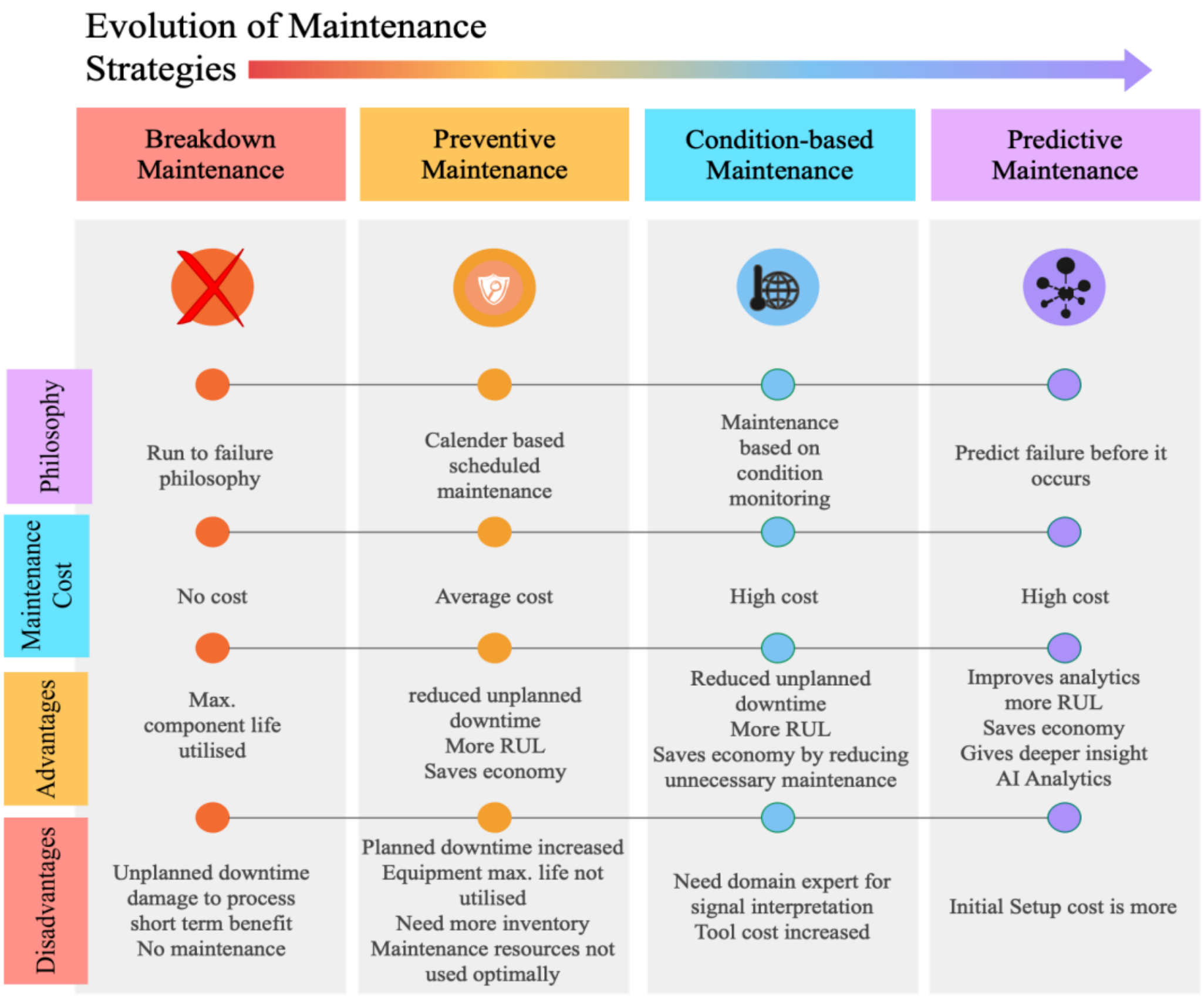

**Figure 4.** Evolution of Maintenance Strategies

## 1.5. Multi-fault Diagnosis in Rotating Machines

Manufacturing process machines, especially the rotating machines, work 24x7, resulting in mishandling or wear and tear due to prolonged use leading to different faults in them. The different components of rotating machines possess different types of faults. These faults are broadly classified at three levels: Component-level faults, System-level faults, and Interrelated faults. Component level faults include the faults related to bearing, shaft, pulley, etc. For example, Bearing is a component that is considered the heart of rotating machines that can have faults like inner race faults, outer race faults, rolling element faults, etc. Similarly, the shaft also is a component that can possess fault like misalignment, rotors in the rotating machines can have unbalance type of fault, and so on. Fig. 5 summarizes different types of component-level faults that can arise the rotating machinery. System-level faults include the faults of the overall system to be monitored. Interrelated faults include the additive faults (addition of different faults due to multiple components, e.g., Unbalance and Misalignment) and multiplicative faults (multiplication of different fault types in a single component, e.g., shaft bent and shaft crack). Additive faults can be better diagnosed as compared to multiplicative faults.

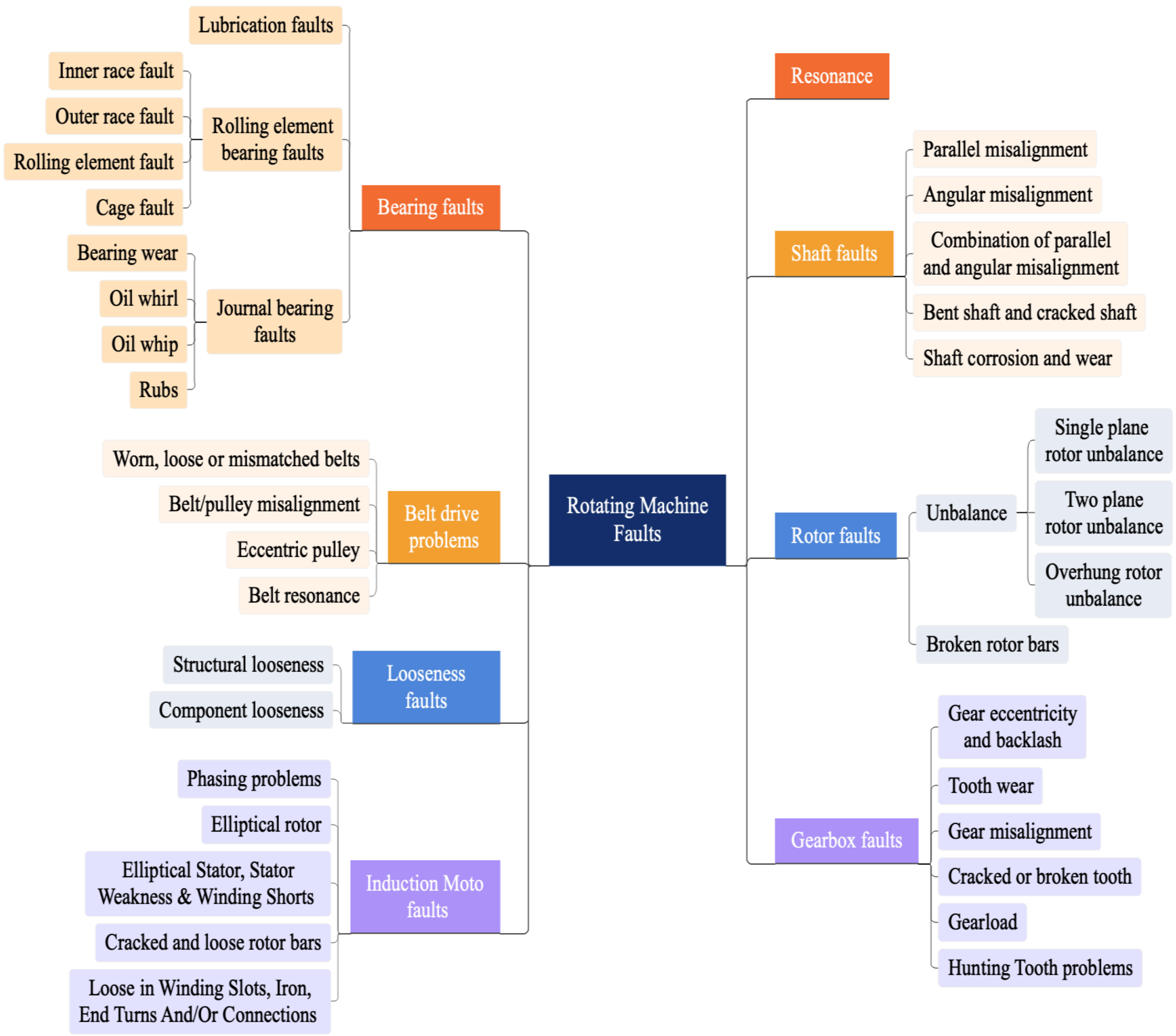

**Figure 5.** General Component-level Faults in Rotating Machines

Many of the completed papers and models have diagnosed a single target fault. In today's era of big data, the single label system ignores the interrelationship of different fault types, making it difficult to accurately determine the location, type, and degree of mechanical failure [47]. Sometimes one fault can give rise to other faults. A combination of different faults is also possible at the same time [18]. Considering this aspect, a proposed model for fault detection is valid or is deployable in real-life situations only if it can detect multiple faults and not just one fault, which is seen in most of the research published. Multi-sensor data fusion [10] also needs to be used to achieve multi-fault diagnosis with high accuracy. The key to accurate fault diagnosis is in identifying and separating the fault characteristics.

## 1.6. Prior Research

The systematic literature review aims at answering the research questions by critically investigating the available research publications. However, as per the articles published related to multi-fault diagnosis of industrial rotating machines, very few papers have done a systematic literature review. Table 1 gives the comparative analysis of different review papers related to fault diagnosis in rotating machines. The table is so formulated that it compares the review papers published with different aspects. The questions are as stated below:

*i.* Are the review papers related to multi-fault diagnosis of rotating machines?
*ii.* Have the authors discussed sensor selection and multi-sensor data fusion?
*iii.* Is there a discussion on where to get data from or how to acquire data?
*iv.* How to extract features and do signal processing?
*v.* What are the different AI Approaches for diagnosis?
*vi.* What are the major challenges and future scope in this field?

A total of five papers have done a systematic literature review related to multi-fault diagnosis of rotating machines [21,22,24,25,28], out of which [21, 24, 25] have discussed multi-sensor data fusion. It is also seen that almost all the papers [21-30] have discussed in detail Artificial Intelligence techniques and models, the challenges in implementing accurate fault diagnosis, and what is the future scope based on the survey. Most of the authors have also discussed the different approaches for fault detection: the statistical approach, the data-driven approach, and the hybrid approach. However, very little importance is given to the equally vital questions like what sensors(s) to use, how to acquire data, how to process the data, with what features to be extracted? All the empty fields in the table also need to be addressed as the proper selection of methods/technology is the key to accurate fault diagnosis. For instance, selecting multiple sensors instead of single sensors increases fault prediction accuracy [10]. Based on this prior research, the authors of this paper have tried to answer all the missed questions and remained unanswered from table 1 to give a comprehensive review related to multi-fault diagnosis in Rotating machines.

Table 1. Comparative Analysis of Review Papers related to Fault / Multi-fault Diagnosis of Industrial Rotating Machines.

| Paper Features / References | Multi fault diagnosis | Rotating machines | Sensor selection | Data source/ test setup | Data acquisition | Feature extraction / processing | Multi-sensor data fusion | AI Techniques | RUL | Challenges | Future Scope | Overview |
|---|---|---|---|---|---|---|---|---|---|---|---|---|
| [21] | ● | ● | | ● | ● | ● | ● | ● | | ● | ● | Role of AI in rotor fault diagnosis |
| [22] | ● | ● | | | | | | ● | | ● | ● | Fault diagnosis in Induction Motor |
| [23] | | ● | | | | | | ● | | ● | ● | Early fault diagnosis in rotating machines |
| [24] | ● | ● | ● | | | | ● | ● | ● | ● | ● | Multidimensional prognostics for rotating machines |
| [25] | ● | ● | | | | ● | ● | ● | ● | ● | ● | diagnostics and prognostics for CBM |
| [26] | | ● | | | | | | ● | ● | ● | ● | Multi-model approach for PDM |
| [27] | | ● | ● | | | | | ● | | ● | ● | PdM & intelligent sensors in smart factory |
| [28] | ● | ● | | | | ● | | ● | | ● | ● | multi-fault diagnosis of planetary gears |
| [29] | | ● | | | | | | ● | | ● | ● | ML Methods for manufacturing process |
| [30] | | ● | | | | | | ● | | ● | ● | Signal Processing methods for fault diagnosis |

## 1.7. Technology focus and evolution time-line for Machine Health Monitoring

All technologies are created with a specific goal in mind. Search engines, for example, were built to sort through the vast amounts of data available on the internet. With each new upgrade, current technologies are combined to create something superior to what was before used. The list goes on and on. It is no surprise that many people have struggled to keep up with the rapid pace of technological advancement. To be fair, the scope of technology is so vast that condensing it all into a single section is impossible. Fig. 6 [20] shows the evolution in technology related to fault diagnosis in industrial rotating machines. The evolution is demonstrated concerning sensor technology, data and signal processing, condition monitoring & diagnosis, and maintenance strategies in the past years and the near future.

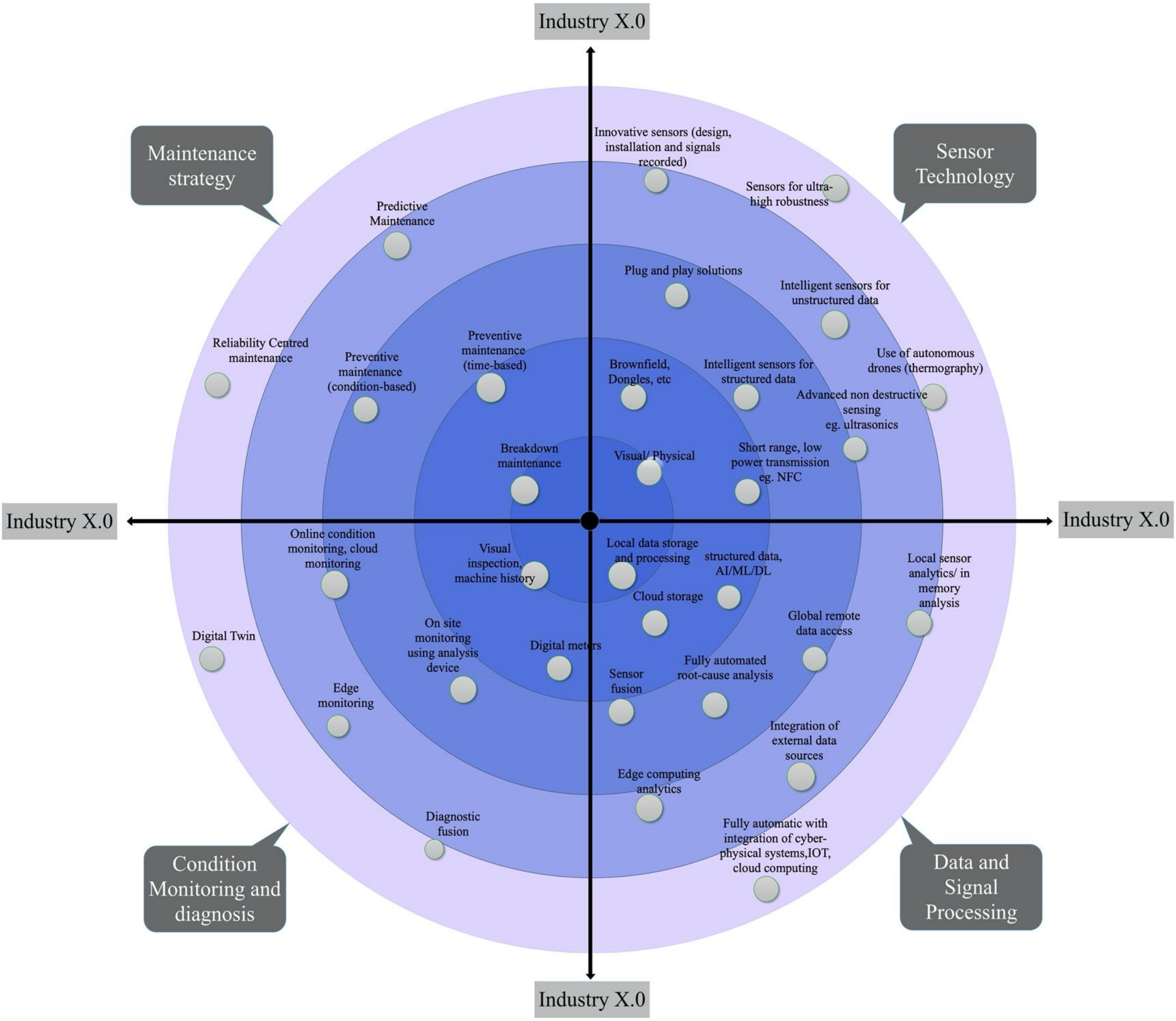

**Figure 6.** Technology Focus and Evolution Time-Line for Machine Health Monitoring.

## 1.8. Research Goal

The presented Systematic literature review critically analyzes existing studies on the multi-fault diagnosis of industrial rotating machines using PRISMA guidelines. The literature review is achieved with the help of the critical questions formulated in table 2. The table explains the research questions along with the discussion related to the question. The authors believe that the answers to the questions in the table will give a solid foundation to the upcoming research in the field.

Table 2. Research Goals

| Sr. No. | Research Question | Discussion |
|---|---|---|
| 1 | What are the different approaches employed to achieve multi-fault diagnosis under Predictive Maintenance (PdM) for rotating machines in Industry? | This section gives a comparative analysis of different PdM approaches discussing the advantages and disadvantages of each approach |
| 1 | What are the available data sources for data-driven PdM? How to select the appropriate sensors for data collection? | Discussion related to online datasets and data collection on test setup is discussed along with Different sensors explanations. |
| 2 | What are the different data acquisition methods? What are the data validation techniques? | Critical points for data acquisition and validation techniques are discussed. |
| 3 | What are the different signal processing techniques? | Feature extraction and the types for signal processing are discussed. |
| 4 | What are the approaches to achieve Information fusion or multi-sensor data fusion? | Multi-sensor data fusion is discussed in detail. |
| 5 | How to implement AI models for multi-fault detection? | A data-driven and a hybrid approach are discussed for PdM. |

## 1.9. Contributions of the study

The SLR's primary purpose is to conduct a critical study of existing current methodologies to implement multi-fault diagnosis in industrial rotating machines to find solutions to the research question posed in Table 1 using PRISMA guidelines. From this analysis, the following are our significant contributions to the field:

The authors have tried to explain all the aspects needed to implement Multi-fault diagnosis in Rotating machines. The paper covers all the significant reviews on data sources from online datasets to building the test setup. The authors have also focused on sensor types and sensor selection. The paper has covered data acquisition details, including the hardware and the software. Signal processing is studied in detail with different features, including the time domain, frequency domain, and time-frequency domain features. The Paper also has briefed the aspects of multi-sensor data fusion, explaining the data level fusion, feature level fusion, and decision level fusion. The data-driven approach for predictive maintenance using different AI algorithms, the platforms to implement them are all systematically covered. Finally, the paper's primary focus is in conclusion, which gives the research gap and prospects that would guide future researchers in the field.

## 1.10. Paper Organization

The paper organization is shown in fig.7. The paper has eight main sections: introduction, research methodology, review results, discussion, Challenges and limitations, future scope, and conclusion. The first section is the introduction explains the rotating machines, the significance of the study, the motivation behind the study, the terminology used, the evolution of maintenance strategy, multi-fault diagnosis, prior research, technology focus and evolution timeline for multi-fault diagnosis, research goal and contribution of the study. Section II is the research methodology that explains the method used for PRISMA's systematic literature review. After the methodology comes the results section III, which is the paper's core that answers significant questions related to the multi-fault diagnosis in industrial rotating machines as formulated in table 2. Next, we discuss the outcome of the survey in section IV, followed by The challenges and limitations in section V. One more important section in the paper is Section VI which gives recommendations for future work with the research gap. Finally, we conclude the paper in Section VII, followed by references.

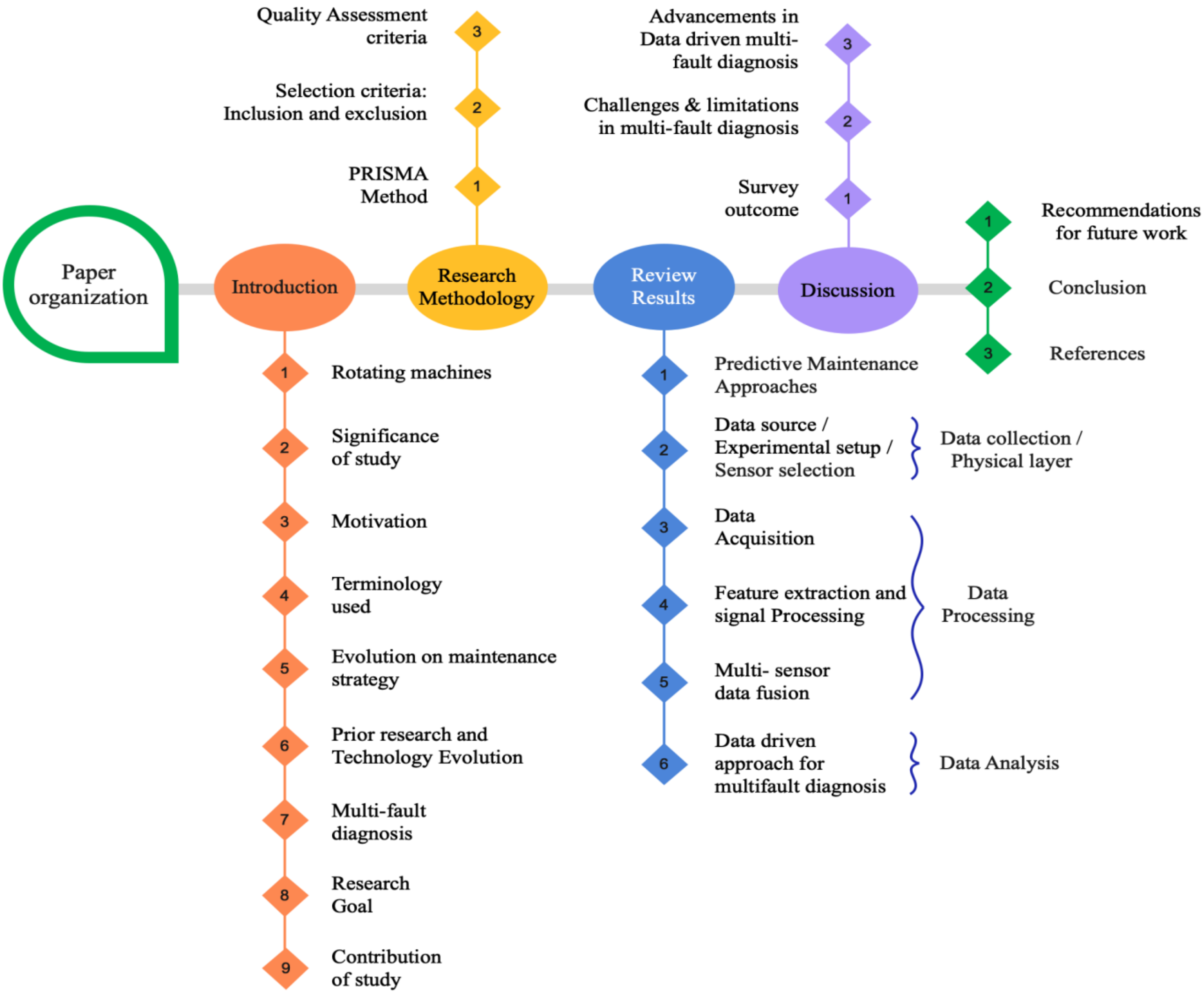

**Figure 7.** Paper Organization

## 2. RESEARCH METHODOLOGY

The procedures or strategies used to find, select, process, and analyze information about a topic are referred to as research methodology. A strategic review is carried out in this paper using the Preferred Reporting Items for Systematic Reviews and Meta-Analysis (PRISMA) guidelines [31,32]. PRISMA is a set of guidelines for the structure and composition of systematic reviews and other data-driven meta-analyses. The systematic review presented in this paper has taken the PRISMA checklist table [32], which has 27 items to be considered while using the PRISMA method for systematic review. This method comprises three steps: framing research questions, the search stage, and the standards for inclusion and exclusion of research papers [31]. The details of these three stages are explained below.

The first stage consists of formulating research questions which were shown in table 2 earlier. The quality of research depends mainly on the research questions framed. Research questions guide us to explore different aspects of the research systematically. The second stage is the search for articles which starts with the identification of a database for articles. In this study, Scopus and Web Of Science (WOS) are used as the database for article selection. In this stage, Search fields are defined concerning the article title, abstract, and keywords. Tables 3 and 4 show the selection procedure for keywords based on the PIOC (Population, Intervention, Outcome, Context) approach published by Kitchenham [33] and the final set of keywords used, respectively.

Table 3. Keyword Selection

| Parameter | Meaning | Keywords used |
|---|---|---|
| Population | It is an Application area | "Rotating Machines" OR "Bearings" |
| Intervention | It is the software methodology | "Artificial Intelligence" OR "Machine learning" OR "Deep Learning" OR "multi-sensor data fusion" OR "Multivariate." |
| Outcome | It should relate to factors of importance to practitioners such as improved reliability, reduced production costs, and reduced time to market | "multi fault diagnosis" OR "Fault diagnosis" OR "multiple faults" OR "Fault detection" |
| Context | It is the context in which the intervention is delivered | "Rotating machines" OR "bearings" |

Table 4. Search Queries on Scopus and WOS with Parameters

| Dataset | Exact query | Result |
|---|---|---|
| Scopus | ("Rotating Machines" OR "Bearings") AND ("multi sensor data fusion" OR "Multivariate" OR "multi fault diagnosis" OR "multiple faults") | 2091 |
| WOS | | 261 |

The third and final stage is to create protocols for assessing the technical and scientific articles that these searches have generated from Scopus and WOS databases to keep only those most relevant articles to the research theme. The third stage is explained in detail in the following subsection, which mainly includes the inclusion criteria, exclusion criteria, quality assessment criteria, and their use for selecting or rejecting the paper.

## 2.1. Inclusion and Exclusion Criteria

A list of inclusion criteria for research paper selection and exclusion criteria for research paper rejection is applied to choose relevant research studies for systematic review. After applying the keyword query search in Scopus and WOS, these criteria are applied to the articles after the second stage. Inclusion and exclusion criteria applied are as shown in table 5.

Table 5. Inclusion and Exclusion Criteria

| Sr. No. | Inclusion Criteria |
|---|---|
| 1. | Articles should be published between 2011 to 2021 |
| 2. | Articles should meet at least one term related to the research theme |
| 3. | Articles should be either published or in the process of publishing in Journals |
| 4. | Articles should give answers to the research questions. Therefore, the title, abstract, and full-text reading are done to achieve this criterion. |
| | **Exclusion Criteria** |
| 1. | Non-English Literature |
| **2.** | Duplicate articles |
| **3.** | Articles with non-availability of full text |
| **4.** | Articles not relevant to Fault Detection in Industrial Rotating machines |

Out of all these criteria mentioned in table 4, the most important criterion is to have the articles related to the research theme. To achieve this, three-stage assessment criteria applied are:

- Abstract-based Assessment: This includes reading the abstract to check whether the article is discussing our research theme. An abstract that matches at least 40% of the research theme is selected for further assessment: the full text-based assessment.

- Full text-based Assessment: This includes a complete reading of the article. Articles that match the research themes are selected, and those that do not match the research theme are rejected. The remaining articles go through the quality assessment.

- Quality-based Assessment: This criterion increases the quality of the literature review. To achieve this, quality assessment criteria are as discussed below.

## 2.2. Quality Assessment Criteria

The quality assessment is based on the following four criteria: The articles that do not meet these criteria are rejected.

C1: Is the paper discussing multi-fault diagnosis in Industrial Rotating Machines?

C2: Is the paper discussing data collection and related topics like sensor selection and sensor data fusion?

C3: Is the paper discussing different feature extraction and signal processing methods?

C4: Is the paper discussing the AI techniques related to fault diagnosis in rotating machines?

The pictorial explanation of the Systematic Literature Review process regarding PRISMA guidelines is shown in fig.8.

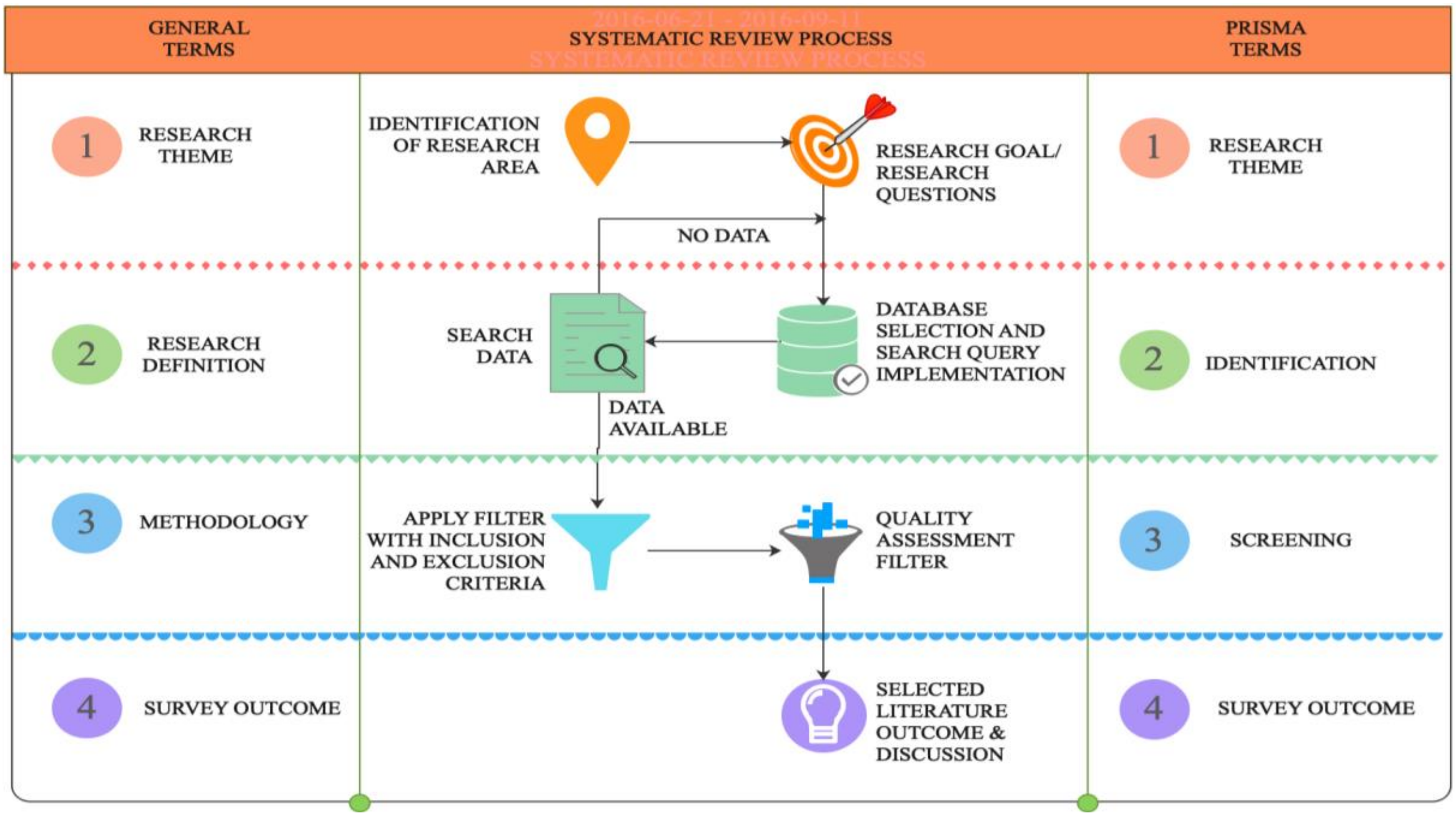

**Figure 8.** Systematic Literature Review Process

### 2.3. Systematic Review Implementation

The following main steps were used to select appropriate papers for this review. The Systematic review flow diagram using PRISMA guidelines, as shown in fig.8, depicts the steps of recognition, screening, eligibility, and inclusion. Figure 9 depicts the systematic search strategy implemented. After formulating the research questions and the keywords for a search query, the next step is to select the database. Scopus and WOS were selected as the database, where initial search results were 2091 articles from Scopus and 261 articles from WOS. The next step was to implement the inclusion and exclusion criteria as discussed in table 5. Accordingly, articles from 2011 to 2021 were selected. Also, articles in the English language were selected related to Engineering and computer science domain areas. Also, a filter was applied to select article-type documents and remove the documents from conference proceedings, review papers, etc. 116 documents from WOS and 1845 documents from Scopus were removed, and 391 articles were selected. The next step was to remove the duplicates. 97 duplicate articles were removed, leaving 294 total articles. The next stage was to apply a filter based on reading the titles of the articles, reading the abstract, and checking the scientific recognition of the articles. 192 misaligned articles were deleted, leading to 102 filtered articles. Also, it is essential to remove unavailable articles (full text). The final step is to read the whole article and discard the misaligned ones. 34 unavailable and misaligned articles were deleted. The final portfolio of articles comprised 68 closely aligned articles. This entire process is shown in Fig.9.

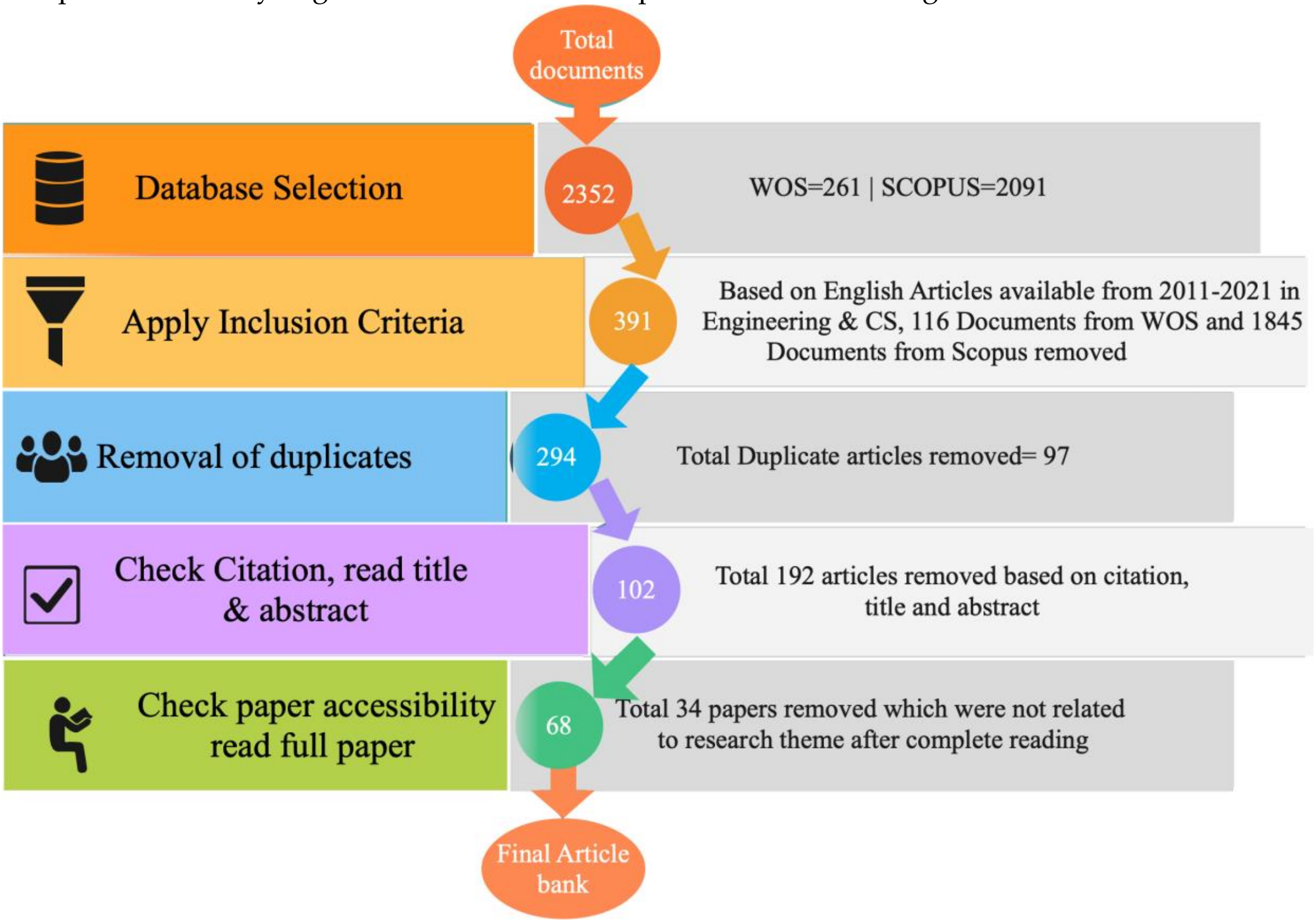

**Figure 9.** Implementation of Systematic Literature Review Process

## 3. REVIEW RESULTS

This section summarizes the findings of our systematic review process. It answers the research questions formulated in table 2 based on the review process results conducted using PRISMA guidelines. Here is the evaluation and summary of the papers.

### 3.1. RQ1. What are the different approaches employed to achieve multi-fault diagnosis under Predictive Maintenance (PdM) for rotating machines in Industry?

Predictive maintenance (PdM) is a type of condition-based maintenance that monitors the condition of assets using sensor devices. These sensor devices supply data in real-time, which is used to predict faults using AI, and intimate when the asset will require maintenance and prevent equipment failure. Specific widely used parameters and methods [1] to detect the fault and predict the condition of Industrial rotating machines are Vibration analysis [15], Motor Current Signature Analysis [13], Temperature analysis, Magnetic chip detectors [16], Infrared Thermography, Acoustic Emission [16], Airborne Ultrasound, Lubrication analysis, etc. A combination of these methods improves the accuracy of analysis to predict the faults. The PdM approach is classified into three major categories: Physics-based Approach, Knowledge-based Approach, and Data-Driven Approach [34,35,36]. These approaches are summarized in fig. 10 [217]. The knowledge-based approach [34,36] combines field experience and computational knowledge from domain experts and sets rules to interpret the faults. The advantage of this approach is that it requires less information and it does not require a mathematical model. However, it is challenging to implement it without historical data and domain expertise. Expert systems and fuzzy logic approaches are two examples of experience-based models that rely heavily on domain knowledge. Physics-based models [46] are methods that use knowledge of a system's failure mechanisms (for example, unbalance growth) to create a mathematical equation model for the system's degradation process. The main advantage of this approach is that it does not require collecting a lot of data, and it can be easily validated. Extrapolation is also easily possible. However, at the same time, it is not suitable for complex processes or machines and requires expert knowledge. Also, considering all degradation mechanisms is a challenging task. These physics-based models include the Finite Element Model (FEM), Kalman filter (KF), and Particle filter (PF), which all rely heavily on mathematical models.

Data-driven models, the most widely used approach, depend highly on big data and analysis. The Analysis allows us to predict the system's condition or state or match comparable examples in the set of past experiences. The advantage of this approach is that it does not require a separate performance degradation process; however, it requires extensive data, and the accuracy depends highly on the algorithm's training. A data-driven approach [37-41] is currently the trend attracting much attention of researchers. It mainly includes Artificial Intelligence models. Using different AI Algorithms, viz., Machine learning, deep learning is employed to build these models. Another effective and emerging approach that anchors the benefits of various currently available prognostics models is the hybrid approach [42-45]. It combines different approaches (knowledge-based, physics-based, and data-driven) that result in a hybrid model with better predicting ability.

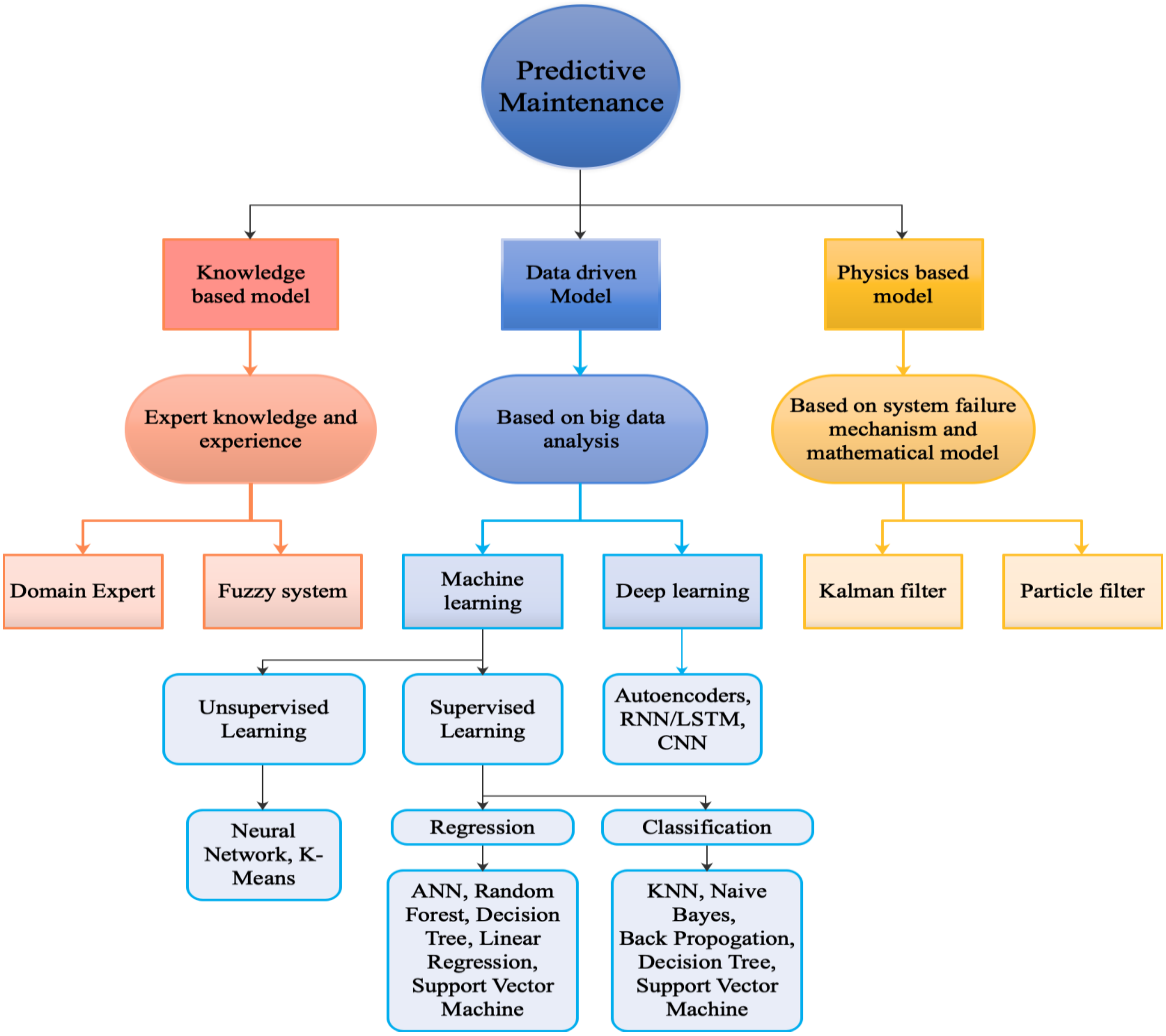

**Figure 10.** Approaches to Implement Predictive Maintenance

### 3.2. RQ2. What are the available data sources? How to select the appropriate sensors for data collection?

Condition monitoring and fault diagnostic research for rotating machines are critical for predictive maintenance, optimal device operation, and workpiece quality. Researchers are concentrating their efforts on two areas to increase this diagnostic accuracy: advanced signal processing technologies and artificial intelligence technology. The dilemma that emerges with the use and development of these new methods and techniques is that they have their benefits and drawbacks and can only be used in specific circumstances. As a result, the hybrid intelligent fault diagnostic technique has been extensively researched in which multiple sophisticated signal processing methods and artificial intelligence approaches are used simultaneously. However, in most diagnostic systems, just one type of information (such as vibration information) is employed [48, 73], resulting in inadequate machinery information, particularly in sophisticated systems. The lack of information even leads to misdiagnosis. Also, the extracted features are focused on a single domain, and features of other domains are ignored. Therefore, the diagnosis model must be based on multi-dimensional and multi-level information or data sources [49]. When we think of Information or data source, there are three main aspects to it. One is the machinery/test setup on which the data is collected, the second is the sensors used for data collection, and the third is how to check the data validity. Let us analyze the different aspects implemented by different researchers.

### 3.2.1. Online Data Source

The data can be achieved in two ways. The first uses an online dataset, and the second uses a test setup to collect data using sensors. There are various online platforms available for online datasets. The most widely used dataset is provided by Case Western Reserve University [50,58]. Ball-bearing test data for both normal and defective bearings are available on this page. Experiments were carried out with a 2 hp Reliance Electric motor, with acceleration data collected close and far from the motor bearings. The precise test settings of the motor and the bearing defect status for each experiment have been meticulously documented on these web pages. Electro-discharge machining was used to seed defects in motor bearings (EDM). Faults with diameters ranging from 0.007 to 0.040 inches were introduced at the inner raceway, rolling element (i.e., ball), and outer raceway individually. The test motor's faulty bearings were replaced, and vibration data was taken for motor loads ranging from 0 to 3 horsepower. The "FEMTO Dataset [66]" is another popular dataset for estimating a bearing's remaining usable life (RUL), which allows researchers to evaluate novel methods for bearing RUL prediction [55]. The FEMTO-ST2 institute designed and built PRONOSTIA, a platform for testing and verifying bearing fault detection, diagnostic, and prognostic techniques. One more dataset is the IMS bearing dataset [56], generated by the NSFI/UCR Center for Intelligent Maintenance Systems (IMS) with support from Rexnord Corp. Unlike other datasets, which intentionally induce bearing defects by scratching or drilling the bearing surface or generating bearing faults by applying a shaft current for accelerated life testing, the IMS dataset provides a comprehensive natural bearing defect history record. The bearing is driven for 30 days in a row at a constant speed of 2,000 rpm for 30 days [56]. Another essential online dataset collection is available at NASA Prognostic Centre of Excellence [51,65]. The Prognostics Data Repository collects data sets donated by universities, government organizations, and businesses. The data repository is dedicated to prognostic data sets, that is, data sets that may be utilized to create prognostic algorithms. Typically, these are time-series data from a nominal state to a failing state. The bearing dataset available on NASA PCoE is provided by the Center for Intelligent Maintenance Systems (IMS) [56], University of Cincinnati. NASA PCoE also has one more bearing dataset that is "FEMTO Dataset [66]" discussed earlier. Another dataset available online is by the Society for Machinery Failure Prevention Technology (MFPT) [59], wherein they provide the Condition Based Maintenance Fault Database whose objective is to give multiple data sets of known good and faulty bearing and gear conditions. This dataset is being made publicly available, along with sample processing code, hoping that researchers and CBM practitioners would enhance the approaches and, as a result, develop CBM systems more quickly. One more platform that provides online datasets is Mendeley data [52,61] which contains vibration signals collected from bearings of different health conditions under time-varying rotational speed conditions. There are 60 datasets in total in this. Next is IEEE DataPort [64], a valuable and easily accessible data platform that enables users to store, search, access, and manage standard datasets. Another bearing dataset [63] from Paderborn University comprises synchronous measurements of motor current and vibration signals, allowing multi-physics models to be verified and sensor fusion of various signals to improve bearing fault detection accuracy. Both stator current and vibration signals are measured with a high resolution, and a high sampling rate and experiments are performed on 26 damaged bearings and 6 undamaged (healthy) ones. Also, one more online dataset is composed of 1951 multivariate time-series acquired by sensors on SpectraQuest's Machinery Fault Simulator (MFS) Alignment-Balance-Vibration (ABVT) [60]. It comprises six different simulated states: normal function, imbalance fault, horizontal and vertical misalignment faults and, inner and outer bearing faults. Table 6 [57] is a summary of all the available online datasets discussed above.

Having access to the online datasets gives monetary benefits as less expense is incurred in developing own test rig. Also, it saves time in making and designing a test setup. However, in the long run, it does not give data freedom. The test setups available online are for fixed conditions and a particular type of fault. If the researchers have to get more data for varied conditions, it is not possible. It can lead to incomplete diagnosis or misdiagnosis. Also, the datasets available online are mostly related to bearing faults. As a result, other significant faults in rotating machines are ignored, and multi-fault diagnosis is impossible. Also, one should check the authenticity of the data before using it in any project.

Table 6. Summary of Available Online Dataset for Fault Diagnosis of Rotating Machines.

| Dataset | Sensor type | No. of sensors | Sampling frequency | Fault generation | Type of fault | Motor Speed | Labeled data | Load |
|---|---|---|---|---|---|---|---|---|
| **CWRU[58]** | Accelerometer | 2 | 12 & 48Khz | Artificial (Single point faults using electro-discharge machining) | Bearing | 1797-1720 | Yes | 0-3 hp |
| **IMS [56]** | Accelerometer | 2 | 20KHz | Natural | Bearing | 2000 | Partially labeled | 6000 lbs |
| **MFPT [59]** | Not specified | - | 97k SPS & 48k SPS | Natural | Bearing | - | Yes | 270 & 300 lbs |
| **MAFAULDA [60]** | Accelerometer, tachometer, microphone | 4, 1 & 1 | 50 kHz | Artificial (Induced unbalance and misalignment) | Bearing, unbalance, misalignm ent | 700-3600 | Yes | - |
| **Mendeley Data [61]** | Accelerometer, encoder for speed | 1&1 | 200kHz | Artificial (not mentioned) | Bearing | - | Yes | - |
| **FEMTO [62] (Pronostia)** | Accelerometer, thermocouple | 2 & 1 | 25.6KHz | Natural | Bearing | - | No | 4000N |
| **Paderborn University [63]** | Accelerometer | 2 | 20 kHz | Natural | Bearing | 1500 & 900 | Yes | 0.7 & 0.1 Nm |
| **IEEE Data Port [64]** | This platform is a collection of datasets of different experiments conducted by different sources. | | | | | | | |
| **NASA PCoE [65]** | | | | | | | | |

### 3.2.2. Offline Data Source using Test Setup

The online datasets, as discussed above, have their advantages as well as disadvantages. However, the biggest drawback is that one cannot study other faults in rotating machines using the available datasets. So though it might be a costly affair, indeed, the best option is to set up a test rig and collect data. Many researchers [73, 67, 68] opted for this option and successfully proved the results. So again, there are two options for the setup: design the test rig [67] or buy the test rig online. Test setups by SpectraQuest, Inc. [69] are widely used by most researchers [68]. Tyrannus Innovative Engineering & Research Academy is another platform that provides machinery fault simulators [70]. Collection of data should be ideally carried out on the real Industrial rotating machines. However, due to industrial protocols and uncontrollable conditions on the site, it is impossible to collect data directly from the Industrial environment. Hence the researchers came up with the machinery fault simulator model [69]. The model may have a different set of combinations of components, but what is shared is that there are two types of components: the driver and the driven type of rotating components that resemble the real-time industrial rotating machine behavior.

### 3.2.3. Sensor types and Selection Criteria

After setting up the test setup, knowing what type of sensors can be used for data collection is crucial. Many sensors are available with different specifications, sensitivity, ranges, functions, applications, etc. Therefore, one needs to correctly choose the sensors considering different factors, as the sensors being the data collectors are the basis of the diagnosis [71]. Sensor selection begins with an awareness of a machine's probable failure modes and the related warning indications. Unbalance, bearing damage, cavitation (pumps), increased machine vibration levels, increased temperature of mechanical parts, loss or decrease of lubricant flow, and cooling water flow are typical warning signals in rotating machinery [73]. Each of these warning signs can be analyzed and monitored using an appropriate sensor [72]. In this part, the sensors most often employed to identify problems in rotating machines [94] at the earliest possible time, notably accelerometers and microphones, are discussed.

- **Accelerometer:** Accelerometers [73], as the name indicates, measure acceleration levels, typically expressed with the sign g (equivalent to gravity's acceleration, 9.81 m/s2). They are installed directly on the surface of (or within) the rotating machine near to bearings. Velocity or displacement may be more critical than absolute acceleration levels for specific applications, although this may be determined by integrating the acceleration data. Accelerometers are widely used in fault detection of rotating machines as they convey considerable information about machinery health [74-76]. A general rise in machine vibration is detected by basic vibration sensors, suggesting a potential machine problem. To detect faults with specific machine components like rolling element bearings or fan blades, more advanced sensors employ FFT (Fast Fourier Transform) signal processing to look at sensor data in the frequency domain [77]. Vibration sensors are classified into several categories, and one should be aware of the distinctions between them. Piezoelectric devices [73] rely on variations in electric current caused by movement. They frequently have a wide frequency response, high sensitivity, and low noise levels, although costly. Because piezoelectric accelerometers have been around for a long time, several types are available for various purposes. Micro-electro-mechanical systems, or MEMS [78], are tiny sensors frequently employed in IoT devices to detect vibrations. These sensors are frequently less expensive than Piezoelectric equivalents. An accelerometer may measure vibration in up to three directions. Movement from side to side, forward/backward, and up/down can all be perceived as vibrations. It always uses a 3-Axis sensor rather than a 1-Axis sensor since they are highly correlated, making it simpler to spot potential failures [79].

- **Eddy current sensor**: Non-contact Eddy-Current sensors [73] indicate a conductive component's position and/or change of position. Magnetic fields are used to operate these sensors. The sensor is equipped with a probe that generates an alternating current at its tip. Eddy currents are tiny currents created by the alternating current in the component we are measuring.

- **Temperature Sensor**: Temperature sensors [81] detect changes in machine conditions in temperature by monitoring essential machine components. RTDs (Resistive Temperature Detectors) and thermocouples are utilized in direct measurement applications. Non-contact infrared sensors are utilized for indirect measuring applications.

- **Oil and Lubricant Sensor**: Particle pollution in lubrication systems is monitored using oil particle sensors [73]. A rise in particle count may indicate that bearings, gearboxes are wearing out. The key benefits of using these sensors are the ability to access lubrication conditions in harsh conditions and on machinery that is not easily accessible. Another advantage is establishing a better predictive and proactive maintenance program to detect the beginning stages of lubricating oil deterioration.

- **Current Sensor**: The current draw of machine components is monitored using current sensors [71]. Monitoring the current draw of a motor is an example of a typical application. The high current drawn over time may indicate motor damage or any other problems arising in other parts of the machine. These sensors are clamped around the motor's electrical wire.

- **Acoustic Emission Sensor**: Acoustic emission sensor [82,97] is a device that transforms a local dynamic material displacement produced by a stress wave to an electrical signal. AE sensors are generally piezoelectric sensors with specific ceramic components such as lead zirconate titanate (PZT) as the main component. They are widely used to detect bearing faults.

Several static, dynamic, and other aspects must be considered when choosing a sensor to measure a physical parameter. Sensor selection changes based on the application area. Let us understand some key points to remember before buying a sensor concerning the multi-fault diagnosis of rotating machines.

First, it is crucial to consider the frequency response of the sensors chosen, especially concerning the accelerometer [96,99]. It is perceived in terms of Hz in a vibration sensor. For example, if a sensor can detect vibrations between 1 and 10 Hz, it is difficult to identify the possible failure at 100 Hz. Table 7 [80,89] summarises the various vibration fault frequencies associated with some common rotating machinery faults. Before buying the sensor, one must check whether their fault frequencies lie in the sensor's frequency range.

Table 7. Fault Frequency and Phase corresponding to Different Faults in Rotating Machines

| **Fault** | **Fault Frequency** (fr=RPM/60) of driver or driven rotating machine | **Phase** | **High vibration Axis** |
|---|---|---|---|
| **Force Unbalance** | fr | $90^0$ (between horizontal and vertical) | Radial (horizontal) |
| **Couple Unbalance** | fr | $180^0$ out of phase (across the bearings) | Radial |
| **Static Unbalance** | | $0^0$ out of phase (across the bearings) | |
| **Dynamic Unbalance** | fr | In between $0^0$ to $180^0$ (across the bearings) | Radial |
| **Overhung Rotor Unbalance** | fr | $90^0$ (between horizontal and vertical)<br>$180^0$ axial phase diff. | Radial and Axial |
| **Angular Misalignment** | fr, 2fr, 3fr | $180^0$ out of phase across the coupling | Axial |
| **Parallel Misalignment** | fr, 2fr, 3fr | $180^0$ out of phase across the coupling | Radial |
| **Misaligned bearing cocked on the shaft** | fr, 2fr, 3fr | $180^0$ out of phase on bearing housing | Axial |
| **Rolling Element bearing** | Rolling element Fault frequency is given by:<br>Outer race defect = $\frac{N}{2} fr\{1-\frac{Rd}{Pd}\cos\alpha\}$<br>Inner race defect = $\frac{N}{2} fr\{1+\frac{Rd}{Pd}\cos\alpha\}$<br>Rolling Element defect = $\frac{Pd}{Rd} fr\{1-(\frac{Pd}{Rd}\cos\alpha)^2\}$<br>Cage Defect Frequency = $\frac{fr}{2}\{1\pm\frac{Rd}{Pd}\cos\alpha\}$<br>(+ sign if the outer race is rotating, - sign if the inner race is rotating<br>N= No. of rolling elements, Fr= shaft rotational speed, Hz, Rd= Rolling element diameter, Pd= Pitch circle diameter, α= Contact angle | --- | Axial |
| **Eccentric rotor** | fr (of motor and driven machine, e.g., Fan) | Either $0^0$ or $180^0$ (between horizontal and vertical) | Radial |
| **Bent Shaft** | fr (if bent near shaft)<br>2fr (if bent near coupling) | $180^0$ axial phase difference | Axial |
| **Mechanical looseness** | fr (structural looseness at machine feet) | $90^0$ or $180^0$ (between horizontal and vertical) | Radial (Vertical) |
| | 0.5fr, fr, 2fr, 3fr (pillow block bolt looseness) | --- | Radial |
| | 0.5fr, fr, 1.5fr, 2fr, 2.5fr,… etc. (improper fit between components) | --- | Radial |
| **Resonance** | When fr=natural frequency of components | Before and after resonance, a shift in phase difference is $180^0$ | Radial |
| **Soft foot** | fr (also can be at 2fr, 3fr, 2 times line frequency..) | --- | Radial (vertical) |
| **Gear faults** | Tooth meshing frequency = Nfr And sidebands at Nfr ± kfr, N: Number of gear teeth, K=1,2,3.. | --- | Radial |

Next is the sensitivity [85] of the sensor. Industrial accelerometers generally have a sensitivity of 10 to 100 mV/g [97], although greater and lower sensitivity options exist [95,96]. A low sensitivity (10 mV/g) sensor is preferred if the machine produces significant amplitude vibrations (more than 10 g RMS) at the measuring location. For, e.g., a 100 mV/g sensor should be utilized if the vibration is less than 10 g RMS. The highest g level should never surpass the sensor's acceleration range. One should also note the temperature range of the sensor [96]. Sensors must be able to withstand the application's temperature extremes. While deciding about the sensor, it is also essential to study the type of data acquisition (DAQ) hardware needed, ignorance of which can be a very costly affair. The sensor interface is crucial because it is the medium to connect to the DAQ. One should also choose between wired and wireless sensors. Wired sensors are always preferred because the data is reliable. Mounting of the wired and wireless sensors is also to be checked. There are four types of mounting, with threaded studs being the best, followed by adhesives, magnets, and probe tips [86]. One more critical factor is the computing technique. Whether edge computing will be implemented or cloud computing will be preferred. Nowadays, the processing is done at the network's edge [87] rather than on cloud servers, which reduces system response time, transmission bandwidth use, cloud storage, and computation resources. This is also an essential factor as a slight error in any decision can lead to monetary losses. Table 8 [72] summarizes sensor characteristics and some of the problems they can detect. The table also includes an overview of the most common faults associated with rotating machines and corresponding sensor requirements. Table 9 gives an analysis of multiple sensors used by researchers in their experimental setup.

### 3.2.4. Challenges concerning Smart Sensors in Industry 4.0:

- The first challenge is related to sensor fusion. Sensor fusion is a technology that combines data from multiple sensors to create a single data point. A sensor fusion algorithm integrates sensor outputs with the highest accuracy and efficiency while consuming the least power and reduced noise. Furthermore, these sensors communicate with an application processor via a sensor hub, and selecting appropriate peripherals for each sensor is critical to the success of sensor-fusion systems.
- The next is the challenge related to security and privacy. Several solutions, both hardware, and software are aimed at resolving privacy and security concerns. The industries would never want their data going out of the organization, especially in cloud computing.
- In a given industry, more than one sensor might be sending data over the same network. As a result, the network traffic increases, resulting in data loss.
- Finally, it is also vital to have an energy-efficient sensor network.

Table 8. Overview of Sensors used in Predictive Maintenance

| Sensor Requirements | Bearing condition | Gear fault | Misalignment | Unbalance |
|---|---|---|---|---|
| Low to medium noise > 100 µg/√Hz | | | ● | ● |
| Low noise <100 µg/√Hz | ● | ● | | |
| Bandwidth : 5x to 10x fundamental frequency | | | ● | ● |
| Bandwidth : > 5 KHz | ● | ● | | |
| Multi-axis sensing | | | ● | ● |
| Low frequency response for slow rotating machines | | | | ● |
| High g-range | ● | | | |

| Measurement | Sensor | Bearing condition | Gear fault | Misalignment | Unbalance | Pump cavitation | Load condition | Heat source / Temperature change | Wear debris | Rotor bar issues | Pressure leaks | Eccentric rotor | Winding issues | frequency | Cost | No. of Axis | Benefits | Challenges |
|---|---|---|---|---|---|---|---|---|---|---|---|---|---|---|---|---|---|---|
| Vibration | Piezo accelerometer | | ● | ● | ● | ● | ● | | | ● | | | | Upto 30kHz | low to high | 1 to 3 | simple design, high frequency response, sensitivity, diagnosis accuracy, high G range | low o/p, Costly, noise issue, expertise needed, low temp. range |
| Vibration | MEMS Accelerometer | ● | ● | ● | ● | ● | ● | | | ● | | | | Upto 20KHz+ | low to high | 1 to 3 | | |
| Sound Pressure | Microphone | ● | ● | ● | ● | ● | ● | | | | | | | Upto 20KHz | low cost | 1 | high frequency response, simple, non invasive, limited fault diagnosis | signal validity, noise issue, low reliability of results |
| Sound Pressure | Ultrasonic microphone | ● | ● | ● | ● | ● | | | | | ● | | | Upto 100KHz | low cost | 1 | | |
| Motor Current | current sensor | ● | | | ● | | | | | ● | | ● | ● | - | low cost | 1 | low noise sensitivity, easy installation, cheap | high freq components lost by filtering |
| Magnetic Field | Hall, magnetometer, search coil | | | | | | | | | ● | | | | Upto 250 Hz | low cost | 1 | position, speed sensor, immune to vibration | less accurate, fluctuates with temp. |
| Temperature | Infrared Thermography | | | | | | | ● | | | | | | - | high cost | 1 | cheap, easy to use, large temperature range, no self heating | less stable, less accurate |
| Temperature | RTD, Thermocouple | | | | | | | ● | | | | | | - | low cost | 1 | | |
| Oil Quality | Particle monitor | | | | | | | | ● | | | | | - | low cost | 1 | simple and effective to prevent bearing failure | costly maintenance, slow operation |

Table 9. Analysis of Multiple Sensors used by Researchers in the Experimental Setup

| References | Fault Type | Decision Making Algorithm | Accelerometer | AE Sensor | Dynamometer | Current sensor | Eddy Current Sensor | Load cell | Tachometer | Pressure Sensor | Laser Speedometer | Optical Encoder |
|---|---|---|---|---|---|---|---|---|---|---|---|---|
| [73] | bearing multi-fault | KNN | ● | | | | | ● | | | | |
| [115] | bearing multi-fault | SVM, Wavelet Packet Transform | ● | | | ● | | | | | | |
| [67] | bearing multi-fault | Mathematical model | | | | | ● | ● | | | | |
| [156] | bearing multi-fault | hierarchical multiscale symbolic dynamic entropy | ● | | | | | | | | | |
| [157] | Rotating machines multi-fault | CNN | ● | | ● | | | | | | | |
| [158] | Multiple bearing and gear faults, rotor crack | SVM | ● | | | | | | | | ● | |
| [159] | gearbox multi-fault | DCNN | ● | ● | | ● | | | | | | ● |
| [160] | Bearing and gearbox | generalized Vold–Kalman filtering (GVKF) + FFT | ● | | | | | | ● | | | |
| [161] | Gearbox multi faults, bearing faults | BSLMS | ● | | | | | | | | | |
| [155] | bearing multi-fault | Multi objective iter- ative optimization algorithm (Morlet wavelet filter+whale optimization algorithm) | ● | | | | | | ● | | | |

## 3.3. RQ3. What are the different data acquisition methods? What are the data validation techniques?

Data acquisition is a critical step in condition monitoring (CM) of rotating machinery [84]. The Data Acquisition System (DAQ) characteristics and the price they represent are significant challenges in their implementation [90]. The analog signal from an accelerometer must be converted to a digital signal and recorded once it has been appropriately conditioned. The data acquisition stage involves gathering measurement data from the sensors and processing the raw signal to extract relevant features that may be used to determine the system's health. In data science, this latter process is referred to as feature engineering discussed in upcoming sections.

### 3.3.1. DAQ Setup Specifications

Following criteria are very important when researching data acquisition hardware for a given application [83].

**Sensor compatibility:** The type of appropriate data acquisition system is typically determined based on the sensor's output. Some questions related to sensor compatibility that generally arise are whether the sensor output is analog or digital, the output range, etc. In addition, the sensor must be compatible with the data acquisition system.

**Sampling Frequency**: According to the Nyquist theorem, the sampling frequency (inverse of the sampling rate) should be at least 2.56 times the maximum frequency present in the signal to catch the vibration signal's profile [83] accurately. Picking a system that can sample at least ten times faster than the fastest signal that needs to be recorded is a valuable piece of advice [97].

**Measurement Resolution:** The resolution is usually provided in bits, converted to acceleration units to compute the resolution. Consider an accelerometer system with a 24-bit resolution, which means it can measure 24 (16,777,216) acceleration levels. DAQ systems usually have a resolution of 16 or 24 bits. Higher the resolution better the data visualization [96].

**Filtering capabilities**: Filtering is an essential function of DAQ used to filter unwanted frequency components or noise. Type of filter used changes from one application to other.

**Dynamic Range of DAQ System**: The variation between the smallest and biggest visible parts of a signal is known as a dynamic range [96]. Systems with greater dynamic ranges improve vertical axis resolution. Most applications can be handled by having a dynamic range higher than 100 dB and a sampling rate of at least 200 kS/s.

**Standalone or PC connection**: PC-connected DAQ systems have the advantage of being smaller as there is no need for separate displays, etc. For the same reason, they are generally less expensive. However, for better data security, a standalone system is preferable.

**Data analysis**: Most DAQ systems integrate built-in data analysis that may be used after collecting the data. If the data is to be analyzed outside the DAQ system, it should be exported in standard formats. The most common analysis platforms are Excel for restricted data sets and Matlab for almost infinite data sets with a large range of built-in analytical capabilities [84].

### 3.3.2. Components of DAQ System

A data acquisition system (DAQ) [84] typically consist of sensors for measuring electrical signals, signal conditioning logic, and other hardware like Analog to digital converter (ADC), multiplexer, ADC, DAC, TTL-IO, high-speed timers, RAM, etc. for receiving analog signals and providing them to a processing system, such as a personal computer. Let us understand the critical components of the DAQ system. After selecting an appropriate sensor, it is essential to implement signal conditioning [83]. Vibration measurement inaccuracy can be caused by insufficient signal conditioning. Signal conditioning is the process of preparing an analog signal produced from a sensor so that it may be measured efficiently and correctly by the data Acquisition system's (DAQ) digitizer. In reality, signal conditioning is one of the essential components of a data collection system since the accuracy of the measurement would be unclear until real-world signals are tailored for the digitizer used by DAQ [100]. Signal conditioning may include filtering, Amplification, isolation, excitation, linearization, etc. An Analog to Digital Converter (ADC) is at the heart of any data acquisition system. This chip receives data from the sensors and transforms it to discrete levels processed by a processor [99].

There are different data acquisition options, including Data loggers, Data Acquisition devices, and Data Acquisition systems. A Data Logger is a self-contained data collection system with an integrated CPU and pre-programmed software. Data loggers are popular because they are portable and easy to use for specialized purposes and may function as stand-alone devices. A data acquisition device [95] (USB, Ethernet, PCI, etc.) has signal conditioning and an ADC, but it cannot operate without being linked to a computer {84]. Plug-in device users can use either preconfigured data acquisition software[99], such as DAQamiTM, or a programming environment, such as PythonTM, C++, DASYLab, MATLAB, and LabVIEW TM. Data collection devices provide a configurable solution for a specific application, with various BUS choices [84] and the ability to utilize your favorite software. Modular data acquisition systems are developed for complicated systems that require several sensor types to be integrated and synchronized. These systems are more challenging to set up and operate, but they are very adaptable. Although modular data acquisition systems are the most expensive choice, many applications require the functionality that only a data acquisition system can provide. The DAQ's last component is a computer that collects all of the data collected by the DAQ hardware for further processing. To make sense of the data, DAQ software is also essential to generate legible and relevant results. To put it another way, the DAQ software serves as a bridge between the user and the DAQ hardware.

### 3.3.3. Types of DAQ Systems

There are different types of DAQ Systems. Wireless DAQ Systems [91], Serial communication DAQ Systems, USB DAQ Systems, and DAQ Plugin boards [92]. Wireless systems consist of wireless transmitters sending data back to a wireless receiver connected to a remote computer. When the measurement needs to be conducted at a distance from the computer, serial communication data acquisition systems are an excellent alternative. There are numerous communication protocols [101], the most popular being RS232, which allows for up to 50 feet of transmission lengths. RS485 outperforms RS232 and allows for transmission lengths of up to 5,000 feet. USB is a new protocol for connecting PCs to data acquisition devices. Because USB interfaces provide power, the data acquisition system only requires one cable to connect to the PC. Data acquisition boards for computers are directly connected to the computer bus. The advantages of utilizing boards include speed (because they are simply linked to the bus) and cost (because the computer provides the cost of packaging and power).

### 3.3.4. Data Validation

Data validation is a critical process that ensures the quality of input data before it is utilized to build models and insights [93]. Whether collecting data in the field, analyzing data, or preparing to deliver data to stakeholders, data validation is crucial for every data handling activity. If the data is not accurate from the start, findings will most likely be inaccurate as well. As a result, data must be verified and validated before being used. The data validation procedure includes several steps:

- Ensure that the data is of the correct type: integer, data, text, Boolean, and so on.
- Checking the value range: minimum and maximum values, as well as correct format (voltage, acceleration, etc.)
- Validating data by applying application-specific requirements such as correct temperature, etc.
- Checking for consistency: for unbalance data logging, horizontal is always higher than axial and vertical, and so on.

Writing a script for data validation may be an option depending on coding language proficiency. Another option is to use software programs designed for data validation. Because AI and machine learning models can only generate legitimate results if constructed with valid data, it is essential to follow all of the procedures outlined above during the data acquisition phase to guarantee that models work with clean data.

## 3.4. RQ4. What are the different signal processing techniques?

Vibration signals recorded using vibration sensors from a rotating machine are frequently in the time domain. Depending on the type of sensor used to acquire the data, they are a collection of time-indexed data points gathered across historical time, signifying acceleration, velocity, or proximity [104]. This data is so extensive and from multiple sources in multiple conditions that it is challenging to inspect and conclude the type of fault visually. As a result, Signal processing is frequently required to clean data and put it into a format from which condition indicators may be extracted, called features. Feature extraction is a part of signal processing that gives the essence of the data. These features are used to distinguish between two different vibration signals. Feature extraction is a dimensionality reduction method that reduces a large collection of raw data into smaller groupings for processing [102]. Feature extraction refers to strategies for selecting and/or combining variables into features to reduce the quantity of data that must be processed while still thoroughly and adequately characterizing the original data set [103]. The speed of learning and generalization phases in the machine learning process is aided by reducing data and the computer's efforts in creating variable combinations (features). The feature extraction is classified under three categories: time-domain features, frequency domain features, and time-frequency domain features [98,220]. Let us understand the different features in detail.

### 3.4.1. Time-domain features

A time-domain graph shows how a signal changes over time. Let us look at statistical functions and other sophisticated approaches that may be used to extract features from time-indexed raw vibration datasets. It will adequately reflect machine health to better comprehend vibration signal processing in the time domain. Different statistical functions are used to extract features from time-domain signals based on the amplitude [104]. Statistical Time domain features [97] are Peak Amplitude, Mean Amplitude, Root Mean Square (RMS) Amplitude, Peak to Peak Amplitude, Crest Factor (CF), Variance and Standard Deviation, Standard Error, Zero Crossing, Wavelength, Willson Amplitude, Slope sign change, Impulse factor, Margin factor, Shape factor, Clearance factor, Skewness, Kurtosis,

Higher-Order Cumulants, Histogram, Entropy [109, 110]. Time Synchronous averaging features include TSA signal, Residual Signal (RES), and Difference Signal (DIFS). Time series regressive models include Auto-Regressive Model (AR Model), MA Model, ARMA Model, ARIMA Model. Filter-based methods include demodulation, the Prony model, and adaptive noise cancellation (ANC). Stochastic parameters such as chaos, correlation dimensions, and thresholding methods, i.e., soft and hard threshold, are considered effective techniques for analyzing time-series vibration signals. Furthermore, the Blind source separation (BSS) is a signal-processing method that recovers the unobserved signals from a set of observations of numerous signal combinations. All these methods and are shown in Fig.11 [105]. Detailed analyses of some of these time-domain features, along with their definition and formula, are given in Table.10 [102, 105].

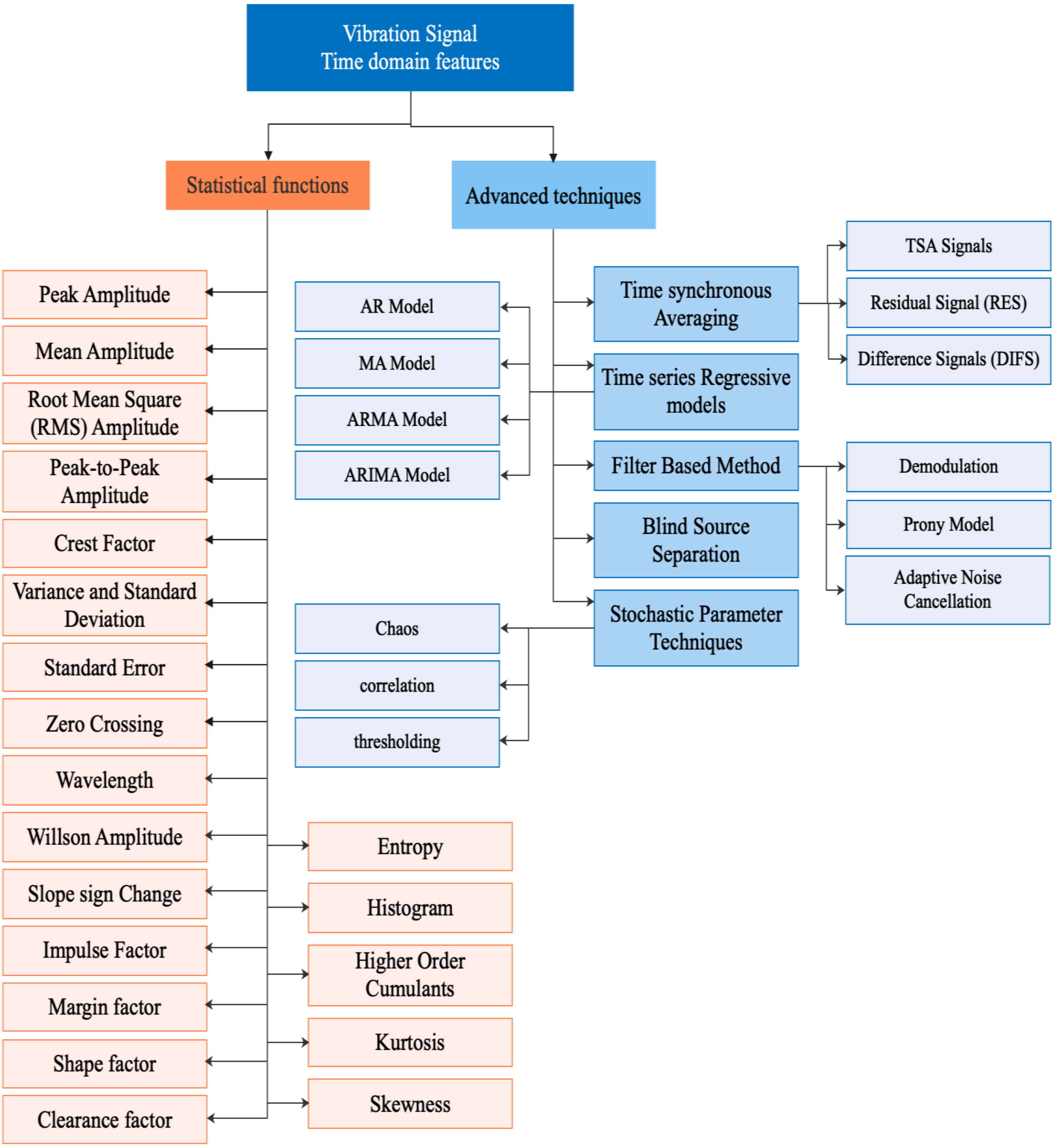

**Figure 11.** Overview of Time Domain Feature Extraction Methods

Table 10. Time Domain Features along with their Definition and Formula.

| Feature | Definition | Formula | Significance / Note |
|---|---|---|---|
| **Peak Amplitude** | Largest positive amplitude of the vibration signal. Indicator of occurrence of impacts. | $x_p = \frac{1}{2}[x_{max}(t) - x_{min}(t)]$ | x=amplitude x(t)=vibration signal. Peak amplitude is valuable for shock events, but it does not consider the time duration or energy in the event. |
| **Mean Amplitude** | Average of the vibration signal throughout a measured interval | $\bar{x}=\frac{1}{T}\int x(t)\,dt$ | T=sampled signal duration |
| **Root Mean Square (RMS) Amplitude** | Variance of the vibration signal magnitude. It is the measure of power contained in the vibration signal | $x_{RMS} = \sqrt{\frac{1}{T}\int \lvert x(t)\rvert^2 dt}$ | It increases as fault develops |
| **Peak-to-Peak Amplitude** | Difference between the highest positive peak amplitude and the highest negative peak amplitude | $x_{p-p} = x_{max}(t) - x_{min}(t)$ | Provides max. excursion of the wave, beneficial wrt. Displacement, specifically clearances. |
| **Crest Factor** | Ratio of the vibration signal's peak amplitude to its RMS amplitude. Faults often first manifest themselves in changes in the peakiness of signal before they manifest in energy represented by RMS. | $x_{CF} = \frac{x_p}{x_{RMS}}$ | Healthy bearing = more CF. Calculates how much impact occurs during the rolling element and raceway contact. CF can provide early warning for faults when they first develop. |
| **Variance and Standard Deviation** | Variance is the deviation of the vibration signal energy from the mean value. The square root of the variance is the standard deviation. | $(V)\ \sigma_X^2 = \frac{\sum(x_i - \bar{x})^2}{N-1}$ <br> (SD) $\sigma_X = \sqrt{\frac{\sum(x_i - \bar{x})^2}{N-1}}$ | N=no. of sampled points <br> $x_i$=element of x |
| **Impulse Factor** | The ratio of the peak value to the average of the absolute value of the vibration signal | $x_{IF} = \frac{x_{peak}}{\frac{1}{N}\sum_{i=1}^{N}\lvert x_i\rvert}$ | Compare the height of a peak to the mean level of the signal. |
| **Clearance Factor** | Ratio of the maximum value of the input vibration signal to the mean square root of the absolute value of the input vibration signal | $x_{CLF} = \frac{x_{max}}{(\frac{1}{N}\sum_{i=1}^{N}\sqrt{\lvert x_i\rvert}\ )^2}$ | This feature is max. for healthy bearings and decreases for the defective ball, outer race, and inner race, respectively. CF has the highest separation ability for defective inner race faults. |
| **Skewness** | The measure of asymmetrical behavior of vibration signal through its probability density function (PDF), i.e., it measures whether vibration signal is skewed to left or right side of distribution of normal state of the vibration signal. | $x_{SK} = \frac{\sum_{i=1}^{N}(x_i - \bar{x})^3}{N\sigma_x^3}$ | $x_{SK}$ For normal condition is zero. <br><br> Faults can impact distribution symmetry and therefore increase the level of skewness. |
| **Kurtosis** | The measure of the peak value of the input vibration signal through its PDF. | $x_{KURT} = \frac{\sum_{i=1}^{N}(x_i - \bar{x})^4}{N\ \sigma_x^4}$ | Good bearing: Kurtosis value <= 3. defective bearing: Kurtosis value > 3 |
| **Histogram** | Discrete PDF of the vibration signal. | LB $= x_{min} - 0.5(\frac{x_{max} - x_{min}}{N-1})$ <br> UB $= x_{max} - 0.5(\frac{x_{max} - x_{min}}{N-1})$ | The lower bound (LB) and upper bound (UB) are two characteristics derived from the histogram. |
| **Entropy** | Measure of the uncertainty of probability distribution of the vibration signal. | $x_{ENT} = \sum_{i=1}^{N} p_{x_i} log\ p_{x_i}$ | $p_{x_i}$=probabilities computed from the distribution of x |
| **AR Model** | Linear regression analysis of the current signal values, i.e., the estimated signal values, of the vibration time series against previous values of the time series | $x_t = a_1 x_{t-1} + a_2 x_{t-2} + \cdots . + a_p x_{t-p} + \mu_t = \mu_t + \sum_{i=1}^{p} a_i x_{t-i}$ | $x_t$=stationary signal, $a_1 - a_p$ are model parameters, $\mu_t$=white noise p=model order |

### 3.4.2. Frequency domain features

One of the most often used vibration analysis techniques for monitoring equipment is frequency analysis, also known as spectrum analysis. In the frequency domain, each sine wave will be represented as a Spectral component. Frequency domain analysis methods, in reality, can reveal information based on frequency features that are difficult to detect in the time domain [105]. As discussed earlier, vibration signals in the time domain are generated by rotating machines. To convert them into the frequency domain, we use Fourier analysis. There are different types of Fourier analysis like Discrete Fourier Transform (DFT), Fast Fourier Transform (FFT), Inverse FFT and DFT, etc. [112], which are used to transform the time-domain signal to the frequency domain. Apart from this, several other approaches for extracting various frequency spectrum features may be utilized to depict a machine's health effectively. The other approaches include Envelope spectrum analysis, power spectrum analysis, spectral kurtosis [108], spectral skewness, Spectral Entropy, Shannon Entropy [113], and some Higher-order spectrum analysis. There are also statistical functions like Arithmetic mean, Geometric mean, Frequency Centre (FC), RMS Frequency (RMSF) [97], Root variance frequency (RVF), median frequency, etc. When a failure occurs, the frequency element changes, and so do the FC, RMSF, and RVF values. The FC and RMSF represent the main frequency position changes, whereas the RVF depicts power spectrum convergence. All these methods and are shown in Fig.12 [102,103,105,106]. Detailed analyses of some of these frequency domain features, along with their definition and formula, are given in Table.11 [102, 103, 105, 106].

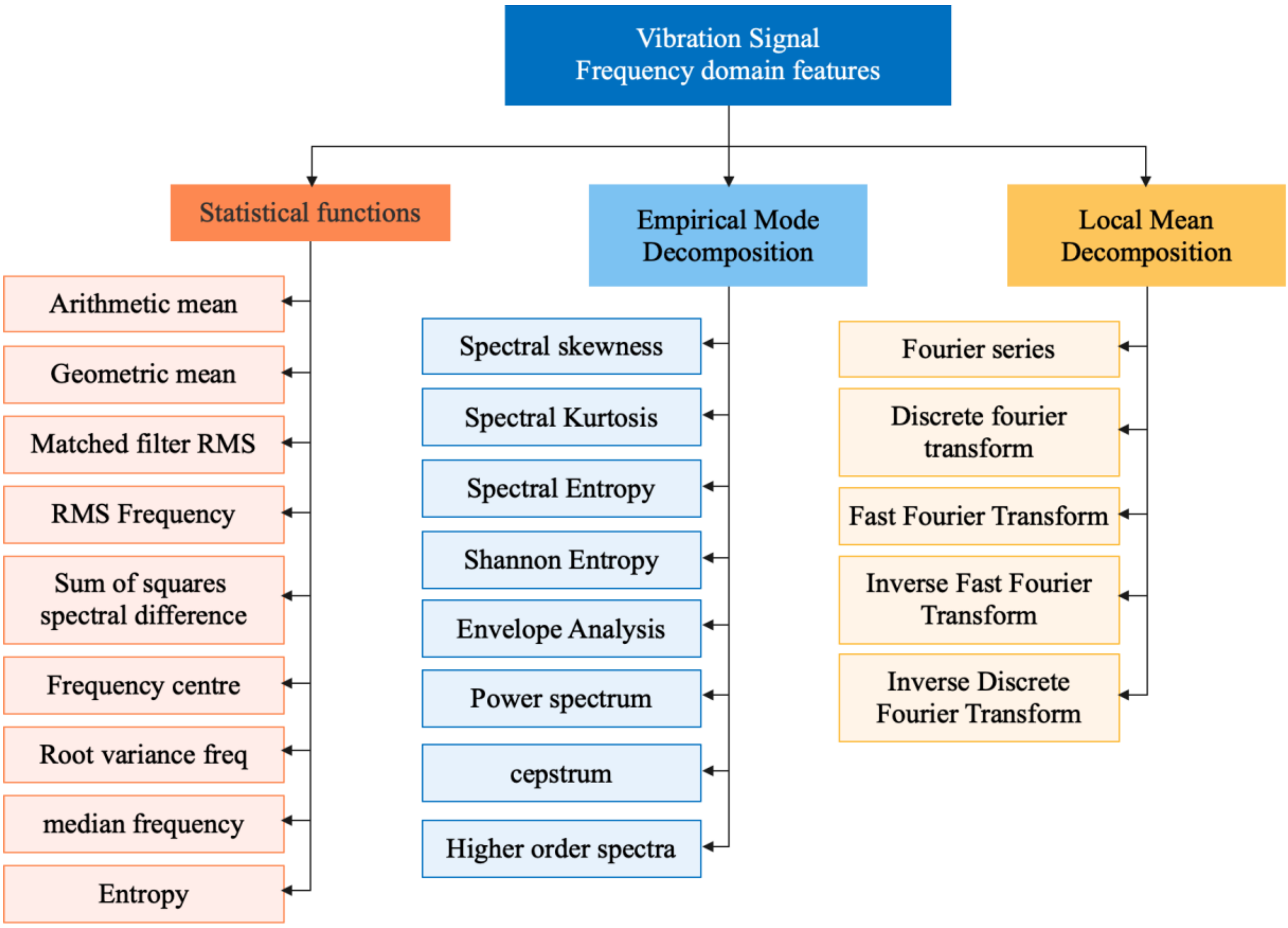

**Figure 12.** Overview of Frequency Domain Feature Extraction Methods.

**Table 11**. Frequency Domain Features along with their Definition and Formula.

| Feature | Property | Formula | Note / Significance |
|---|---|---|---|
| **Discrete Fourier Transform** | Used to convert time domain signal to the frequency domain | $x(w) = \int_{-\infty}^{\infty} x(t)e^{-jwt}dt$ | w=2π/T<br><br>It can be used to find the root of the fault. |
| **Frequency Centre** | Indicate the position changes of main frequencies | $FC = \frac{\sum_{i=2}^{N} x_i' x_i}{2\pi \sum_{i=1}^{N} x_i^2}$ | --- |
| **RMS Frequency** | Indicate the position changes of main frequencies | RMSF=$\sqrt{MSF}$ | $MSF = \frac{\sum_{i=2}^{N}(x_i')^2}{4\pi^2 \sum_{i=1}^{N} x_i^2}$<br><br>Represents overall level of energy across freq. range |
| **Root Variance Frequency** | Shows the convergence of the power spectrum | RVF=$\sqrt{MSF - FC^2}$ | --- |
| **Spectral Skewness** | SS measures the symmetry of the distribution of spectral magnitude values around its mean. | SS(n)=$\frac{2\sum_{k=0}^{\frac{B_L}{2}-1}(\lvert X(k,n)\rvert - \mu\lvert X\rvert)^3}{B_L.\sigma_{\lvert X\rvert}^3}$ | --- |
| **Spectral Kurtosis** | Measures the distribution of the spectral magnitude values and compares them to a Gaussian distribution | SK(n)=$\frac{2\sum_{k=0}^{\frac{B_L}{2}-1}(\lvert X(k,n)\rvert - \mu\lvert X\rvert)^4}{B_L.\sigma_{\lvert X\rvert}^4} - 3$ | --- |

### 3.4.3. Time-frequency domain features

Rotating machines, in general, generate stationary vibration signals. However, some rotating machine analysis is focused on analyzing vibrations during a speed change. As a result, nonstationary signals with changing frequency content are common. When we convert a signal to its frequency domain, we assume that its frequency components do not vary over time, i.e., the signal is stationary. As a result, the Fourier transform in the frequency domain cannot offer information on the time distribution of spectral components. Therefore time-frequency domain needs to be used for nonstationary waveform signals, which are very common when there is machinery failure [105]. Standard time-frequency domain analysis techniques include Short-time Fourier transform, Wavelet analysis, Empirical mode decomposition [114], Hilbert-Huang Transform, Wigner-Ville distribution, Local Mean Decomposition, Kurtosis, and Kurtogram [111,115], etc. These techniques convert one-dimensional time-domain signals into a two-dimensional time-frequency function [107]. All these methods and are shown in Fig.13 [102, 105]. Detailed analyses of some of these frequency domain features, along with their definition and formula, are given in Table.12 [102, 105, 107]. Also, the literature analysis of some of the articles related to feature evaluation is shown in Table 13.

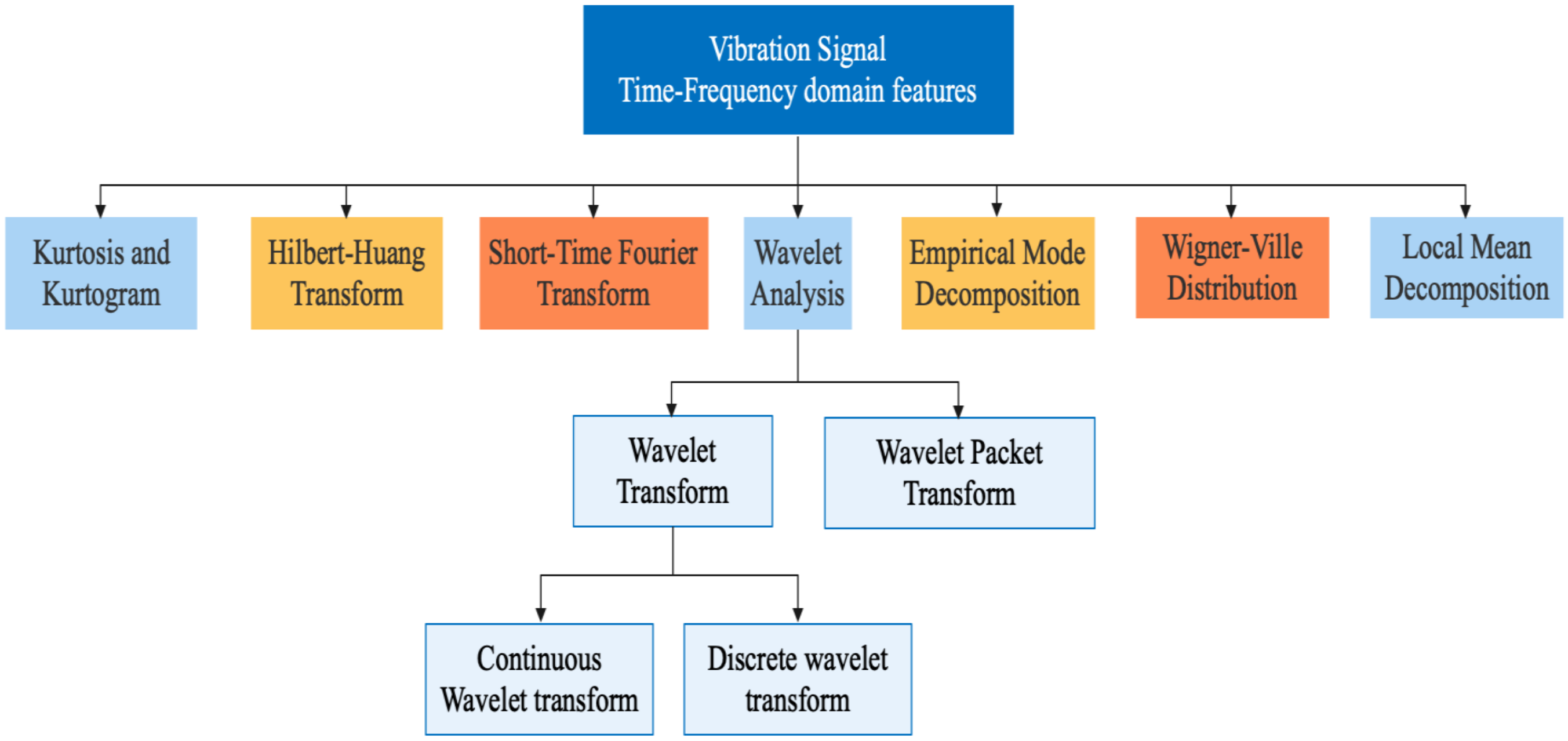

**Figure 13.** Overview of Time-Frequency Domain Feature Extraction Methods.

**Table 12**. Time-Frequency Domain Features along with their Definition and Formula.

| Feature | Property | Formula | Note |
|---|---|---|---|
| **Short-Time Fourier Transform (STFT)** | Instead of computing the DFT of the whole signal, we decompose a signal into shorter segments of equal length | $STFT_{x(t)}(\mathrm{t,w}) = \int_{-\infty}^{\infty} \mathrm{x(t)w(t-\tau)exp(-jwt)d\tau}$ | $\tau$ = time variable<br>$\mathrm{w}(\tau)$ = window function |
| **Continuous Wavelet Transform** | Extension of the Fourier transform maps the original signal from the time domain to the time-frequency domain. | $W_{x(t)}(s,\tau) = \frac{1}{\sqrt{s}} \int x(t)\varphi^*(\frac{t-\tau}{s})dt$ | $\varphi^*$ is the complex conjugate of $\varphi(t)$ |
| **Discrete Wavelet Transform** | Used for the computerized implementation and analysis process of wavelet transforms. | $W_{x(t)}(s,\tau) = \frac{1}{\sqrt{2^j}}\int x(t)\varphi^*(\frac{t-k2^j}{2^j})dt$ | J and k are integers |
| **Wavelet Packet Transform** | DWT is further decomposed into an approximation signal and a detail signal | $d_{j+1,2n} = \sum_m h(m-2k)d_{j,n}$ | m is no. of coefficients and $d_{j,n}, d_{j+1,2n} and\ d_{j+1,2n+1}$ are wavelet coefficients at sub-bands n, 2n, 2n+1 resp. |
| **Empirical Mode Decomposition** | Nonlinear and adaptive data analysis technique that decomposes the time domain signal x(t) into different scales or Intrinsic mode functions(IMF) | $x(t) = \sum_{j=1}^{n} c_j + r_n$ | $c_j\ is\ jth\ IMF\ and\ r_n$ Is residual of data x(t) after extraction of n IMFs. |
| **Local Mean Decomposition** | Decomposes a complicated signal into a set of product functions (PFs), each of which is the product of an envelope signal and a purely frequency modulated signal. | $x(t) = \sum_{i=1}^{m} PF_i(t) + u_m(t)$ | m is the number of PFs. |
| **Frequency Domain Kurtosis** | The ratio of the expected value of the fourth-order central moment of STFT to the expected value of the square of the second-order central moment of STFT | $\boldsymbol{x_{FDK}(f) = \frac{E\{[x(q,f_p]^4\}}{E\{[(x(q,f))^2]^2\}}}$ | $\boldsymbol{x(q,f) = \sqrt{\frac{h}{m}}\sum_{i=0}^{m-1} w_i x(i,q) exp(-jf_p i)}$ where h=interval between successive observations of the process; $w_i = 1$; f=2$\pi/m$; q=1,2,…n ; p=0,1,2,…m ; j=$\sqrt{-1}$; x(i,q) is i/p signal |

**Table 13.** Literature Analysis related to Features Studied

| PAPER | Time domain features | | | | | | | | | | | | Frequency domain features | | | | | | Time-frequency Domain | |
|---|---|---|---|---|---|---|---|---|---|---|---|---|---|---|---|---|---|---|---|---|
| | Mean | RMS | Variance | Peak | Kurtosis | Peak to peak | Skewness | Shape factor | Crest Factor | Impulse factor | Clearance factor | Standard Deviation | RMS Variance frequency | Root Variance frequency | Frequency centre | Mean square frequency | Spectral Skewness | Spectral Kurtosis | Wavelet Energy | Wavelet Packet Decomposition |
| [121] | ● | ● | ● | ● | ● | ● | ● | ● | ● | ● | ● | | ● | ● | | | | | | |
| [122] | ● | ● | | | ● | | ● | ● | ● | ● | | | | | | | | | | |
| [123] | ● | | | | | ● | | | | | | ● | | | | | | | | |
| [124] | | ● | | | ● | | | | ● | | | | | | | | | | | |
| [125] | ● | | | | ● | | | | | | | ● | | | | | | | | |
| [126] | | ● | | ● | ● | | ● | | | | | | | | ● | ● | | | | |
| [127] | | ● | | | ● | | ● | | ● | | | ● | | | | | | | | |
| [128] | | ● | ● | | ● | ● | ● | | | | | | | | | | ● | ● | ● | |
| [115] | | | | | | | | | | | | | | | | | | | | ● |
| [130] | ● | | | | ● | | ● | ● | | | | | | | | | | | | |

### 3.5. RQ.5. What are the approaches to achieve Information fusion or multi-sensor data fusion?

Predictive maintenance has currently been the trend of Industry 4.0, where the machine is monitored for early detection of the fault and avoid failure before it occurs. A single sensor focuses on one parameter avoiding the broader aspects of data, thereby degrading the data quality and likely to induce error in monitoring complex equipment. Hence, multiple sensor configuration technology is introduced to collect extensive information of a machine to enhance monitoring capabilities in terms of measurement accuracy and data richness to improve precision, resolution, efficiency, robustness, and reliability of the entire system. However, complex data collected by multiple sensors create a challenge for data integration and analysis. Therefore, Multisensor data fusion techniques are in great demand for future applications [116]. The advantage of multi-sensor data fusion is that sensor data collected from several locations span a broader display range of the actual situation at the machinery. Also, the complete examination of each sensor's data complements the information while reducing discrepancies. It also increases the data credibility [111]. Information fusion technology can be divided into three fusion methods: data-level fusion, feature-level fusion, and decision-level fusion. Each fusion method has its advantage and disadvantage. The choice of fusion method is determined by the application of the user and types of sensors. In the data-level fusion approach, data from commensurate sensors are fused directly. The fused data is then subjected to feature extraction, followed by a pattern recognition method for classification. The models for data fusion include a weighted average method, Kalman filter method, mathematical statistics method, etc. At this level, fused data is more accurate and trustworthy than data obtained from a single sensor. The sensors employed at this level, however, must be comparable. That is, the measurement must be the same or have physical qualities or occurrences that are identical. The weighted value of multi-sensor signals is difficult to establish when using the most prevalent data-level fusion approach, such as weighted fusion. As a result, data-level applications are constrained in the real world. Data-level information fusion is as shown in fig. 14[194, 119].

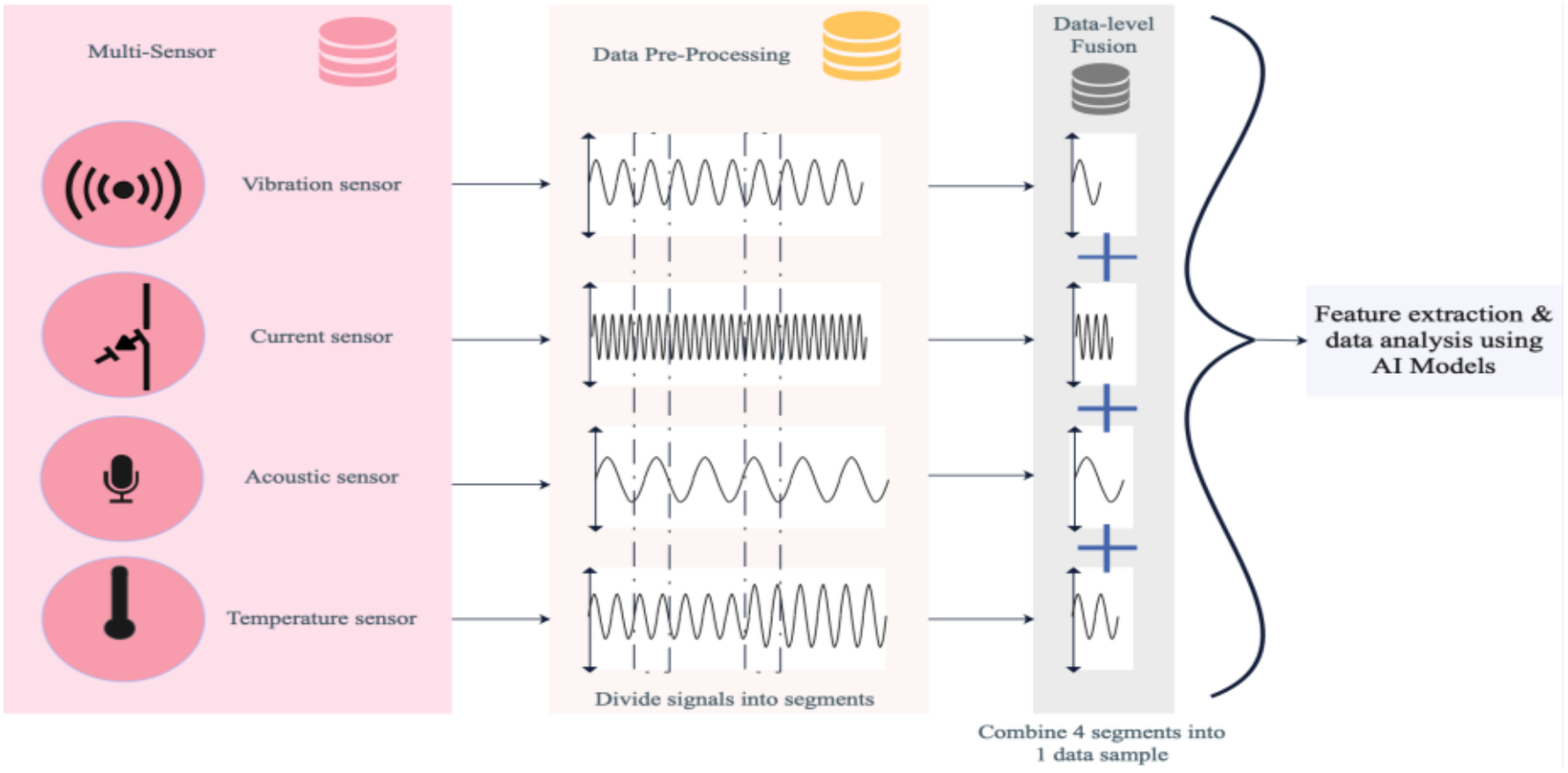

**Figure 14.** Data-level Information fusion.

In the feature-level fusion, each sensor is used to collect a kind of signal. Feature extraction is subsequently applied to obtain a feature vector. All features are combined to identify the best subset of features, then input into a classifier or decision-level fusion. All non-commensurate sensor features are merged to form a larger single feature set, subsequently employed in a specific classification model. Here, heterogeneous sensors are allowed, and information compression is still superior. As a result, feature-level applications are versatile and widely used. The integration of feature-level algorithms mainly includes the Kalman filter method, fuzzy inference method, neural network method, etc. Decision-level fusion is a type of high-level fusion that includes combining the results of several sensors. It stresses the necessity of combining classifiers for a better outcome. The decision level fusion helps minimize the misdiagnosis rates in both false positives and false negatives [73]. At decision-level fusion, features calculation and pattern recognition processes are applied in sequence for single-source data obtained from each sensor. Decision-level fusion methods are then utilized to fuse the decision vectors. The integration of the decision layer mainly includes Bayesian reasoning, D-S Reasoning, etc. A brief overview of feature fusion and decision level fusion is shown in fig. 15 [195, 199].

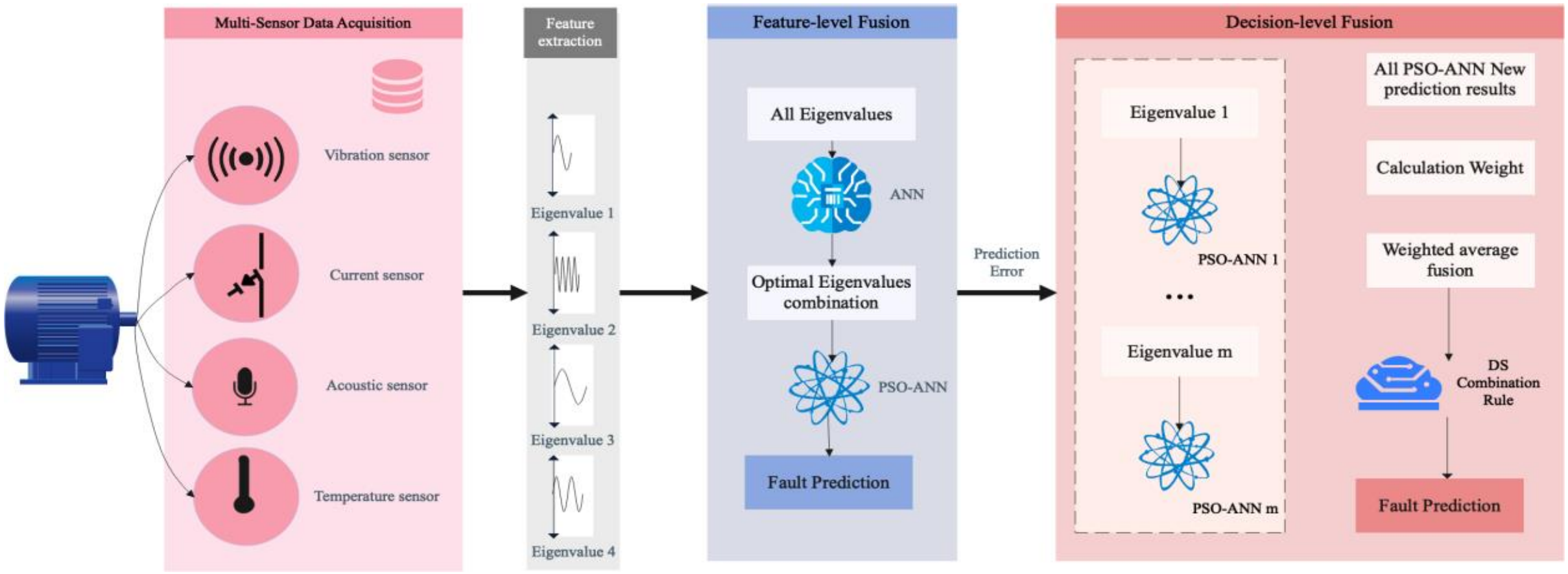

**Figure 15.** Information Fusion at Feature-Level and Decision-Level.

Based on the type of sensors used, multi-sensor fusion is of three types [130]. Global multi-sensor fusion where homogenous processing data given by many sensors but represented in a single reference frame is required. The available data is then analyzed as if it were a single source of information. Second is local/global multi-sensor fusion, where data acquired from different heterogeneous sources is processed. Each sensor's data will be handled individually in this scenario, and the resulting output information will be homogenous. This information homogeneity may then be used to execute a global fusion step. Last is the hybrid multi-sensor fusion, which compromises both global multi-sensor fusion and global-local multi-sensor fusion. This design prioritizes highly reliable sensors by providing a local treatment to their data to reduce the impact of less capable sensors. In addition, this design allows for the collection of all homogeneous sensor data to improve output accuracy.

### 3.5.1. Challenges in multi-sensor data fusion:

There are a variety of factors that make data fusion difficult, stated as follows:

- The bulk of difficulties stem from the data to be merged, the inaccuracy and diversity of sensor technology, and the application environment [117].
- The data collected by sensors is always subject to some degree of imprecision and uncertainty in the readings. Data fusion techniques should be able to convey such flaws [216] adequately.
- Fusion systems should also handle homogeneous and heterogeneous data as the data can be from a similar type of multiple sensors or multiple types of multiple sensors [118].
- Measurements acquired throughout the monitoring process in multi-sensor fall detection are generally unrelated and come from diverse sources. Not only should this data be combined in the most efficient way possible, but they should also be examined for common patterns and commonalities [218].
- Various data sections may travel different paths before reaching the fusion center in distributed fusion setups, resulting in data arriving out of order. To minimize performance deterioration, this issue must be managed appropriately, especially in real-time applications.

- Multi-sensor fusion, as previously said, aids in fault diagnostic reliability. Simultaneously, more sensor devices might considerably raise the monitoring framework's final cost. One should also consider the data acquisition system, which may be costly due to multiple sensors.
- Finally, It is also essential to decide the way of analysis: edge computing or cloud computing. Cloud computing is again very complex due to large data over the cloud [120].

## 3.6. RQ6. How to implement AI models for multi-fault detection?

Artificial Intelligence (AI) has become a widely used technique for data-driven fault diagnosis [131]. The three primary duties of fault diagnosis are determining if the equipment is normal or not, locating the incipient failure and its cause, and forecasting the fault development trend. Artificial intelligence (AI) has garnered much interest from researchers to accomplish these three objectives, and it shows promise in rotating machinery defect identification applications [135]. Because of their resilience and adaptability, AI algorithms for fault diagnostics in rotating machinery have grown widespread.

AI-based methods are classified into two categories, considering the profundity of model structures, i.e., Machine learning (ML) and Deep learning (DL). Machine learning is a technique for processing data, learning from it, and making decisions based on what has been learned [132]. Deep learning creates an "artificial neural network" that can learn and make smart decisions by stacking algorithms. In other words, Deep learning can be called a subset of machine learning. Next, based on supervision, Machine learning is classified as supervised and unsupervised learning. Algorithms like K-Means [133], Neural network (NN), etc., are examples of unsupervised ML. Supervised machine learning [134] is of two types based on the tasks to be accomplished. They are classification and Regression. Classification algorithms are used to classify multiple faults in machinery. Regression algorithms assist us in predicting the upcoming faults based on experience or, in other words, the Remaining Useful Life (RUL) of machinery. K-Nearest Neighbor (KNN)[221], Back Propagation (BP), Support Vector Machine (SVM) [135,173, 222] are examples of classification algorithms. Random forest (RF), Artificial Neural network (ANN) [136], Linear regression (LR), Gaussian Process Regression (GPR), Support vector regression (SVR) are some algorithms for regression [137]. Deep learning can be a comprehensive and effective response to most questions related to machine health monitoring systems (MHMS) for addressing big data and learning multi-scale/ multi-facet/ stratified representation. There are different variants of Deep learning algorithms such as Recurrent Neural Networks (RNN) [88], Auto-encoders (AE) [121], Convolutional Neural Networks (CNN) [88], Deep Belief Network (DBN) [121,222], and Deep Boltzmann Machines (DBM) [88]. In recent years, researchers have successfully implemented these models in the field of Predictive maintenance. Sometimes hybrid of these methods, that is, multiple data-driven algorithms or a combination of data-driven and physics-based algorithms, are used, which are more effective.

A representation of different algorithms used for fault diagnosis and the types of faults diagnosed using both data-driven and hybrid approaches are depicted in Table 14 and Table 15. Table 14 analyses the data-driven approaches currently employed by researchers for fault diagnosis in industrial rotating machines, as discussed earlier. Table 15 is the analysis of hybrid approaches using knowledge-based, physics-based, and data-driven approaches used by the researchers. For example, B et al. (2020) [139] presented a hybrid knowledge-based and data-driven approach for fault diagnosis using Fuzzy logic and a Neural network to predict RUL. D et al. (2001) [140] proposed a hybrid of knowledge-based and physics-based approaches using Fuzzy logic and Kalman filter to predict damage in aircraft actuator faults. Zhang et al. [141] proposed a similar hybrid of fuzzy logic and wavelet neural networks to calculate RUL. Liu et al. (2012) [43] proposed a hybrid, multiple data-driven approach using HMM and SVR to predict the RUL of bearings again. Celaya et al. (2011) [45] proposed a hybrid data-driven and physics-based approach using GPR and KF/PF to predict the RUL of power devices. The experimental results from all these studies show that the hybrid approach yields better results than other independent approaches. Hybrid approaches boost the advantages and conceal the disadvantages of different independent approaches to yield a better diagnosis. However, there are a few challenges while employing the hybrid approach. This includes which kind of approach is appropriate, availability of data and other resources, proper fusion technique of the data as the data can be of various types, with different sampling rates, etc.

Table 14. Contribution of Data-driven Approach for Fault Diagnosis of Rotating Machine

| Algorithm | Algorithm type | Definition | Diagnosis Application | Accuracy | Advantages | Limitations | Ref. |
|---|---|---|---|---|---|---|---|
| **Deep belief network (DBN)** | Deep Learning | The DBN is a deep architecture with multiple hidden layers that can learn hierarchical representations automatically in an unsupervised way and perform classification at the same time. | Fault in Induction motor, Bearing faults, etc. | 99.9% [164] 97.82% [121] | Efficient usage of hidden layers processes unlabelled data effectively, and the problem of overfitting and underfitting may also be avoided | Increasing run time complexity. | [142] [164] [121] |
| **Auto Encoder (AE)** | Deep Learning | An AE is an unsupervised neural network with three layers, which could take an input vector to form a high-level concept in the next layer through a nonlinear mapping. | Bearing faults | 100% [170] 98.3% [171] | Can combine multi-sensor data, prior knowledge not essential | Wide range of data needed, unable to identify relevant information | [121] [170] |
| **Deep Neural Network (DNN)** | Deep Learning | DNNs have deep architectures containing multiple hidden layers, and each hidden layer conducts a non-linear transformation from the previous layer to the next one. | Fault in Bearings and gearbox | 99.06% [167] 100% [112] | automatically deduced and optimally tuned for the desired outcome, robust, flexible | Needs wide range of data for training, overfitting issue | [144] [167] [112] |
| **Convolutional Neural Network (CNN)** | Deep Learning | A CNN is a Deep Learning algorithm that can take in an input image, assign importance (weights) to various aspects/objects in the image and be able to differentiate one from other | Fault in bearings, unbalance, motor faults, pump faults | 93.54% [165] 90.24% [169] | Needs less storage, is less complex compared to ANN, better autodetection | Needs wide range of data for training, overfitting issue, the high computational cost | [145] [165] [169] |
| **Artificial Neural Network (ANN)** | Deep Learning | An ANN is based on a collection of connected units or nodes called artificial neurons, which loosely model the neurons in a biological brain. | Fault in bearing, misalignment, rotor bar faults | 94% [136] 97.89% [102] | Good adaptability, high tolerance for defects, better prediction, and classification. | Needs wide range of training data, overfitting issue, std. network structure not defined | [135] [136] [102] |
| **RNN/LSTM** | Deep Learning | A recurrent neural network (RNN) is a class of ANN where connections between nodes form a directed graph along a temporal sequence. | Gearbox fault, tool fault, RUL, bearing fault | RNN: 94.6% | RNN: easy to process for long i/p, weight can be shared across time-steps. LSTM: better for time-series | RNN: more computation time, hard for training, gradient vanishing problem. LSTM: Requires more | [128] [138] |
| | | | | LSTM: 99.2% [138] | data, can deal with vanishing gradient problem. | time and resources for training. | |
| **KNN** | Machine Learning | KNN is an algorithm where k denotes the number of nearest neighbors used to predict the class of a sample. | Faults like Unbalance, bearing faults | 98.61% [157] | No training period, simple, new data can be added easily | Not suitable for large dataset, sensitive to noisy data, need feature scaling | [143] [157] |
| **SVM** | Machine Learning | It provides a boundary in the input feature space by considering patterns and relationships on the inputs. Boundaries are then used to create a decision rule used for classification | Faults like Unbalance, bearing, misalignment, wind turbine | 98.83% [157] 80% [172] | Better performance for adequate sample size, semi and unstructured data, high dimensional data | Depends on the penalty parameter which is selected by the trial and error method, a standard kernel function is not defined | [143] [157] [172] |
| **Random forest** | Machine Learning | RF is an ensemble learning method for classification, a regression that operates by constructing a multitude of **decision** trees at training time and outputting the class that is the mode of the classes (classification) or mean/average prediction (regression) of the individual trees | Faults like unbalance in the rotor, bearing, impeller, blower | 87.2% [172] | Reduces overfitting, flexible, it automates missing value present in data | Need high computational power, time for training suffers interpretability and fails to determine the significance of each variable | [146] [172] |
| **Extreme Learning Machine (ELM)** | Machine Learning | ELM is a feedforward neural network for classification, clustering, regression, with a single layer or multiple layers of the hidden node | Fault in bearings, gearbox | 98.84% [138] 95.40% [166] | Short training time, simple to use | Overfitting | [138] [166] |

Table 15. Contribution of Hybrid PdM Approach for Fault Diagnosis of Rotating Machine

| Algorithm | Hybrid PdM Approach | Definition | Diagnosis Application | Ref. |
|---|---|---|---|---|
| **Fuzzy logic +NN** | Knowledge-based + Data-Driven | Fuzzy Logic is an approach to variable processing that allows for multiple values to be processed through the same variable | Gearbox faults, RUL Prediction | [139] |
| **Fuzzy logic + Wavelet Neural Network** | Knowledge-based + Data-Driven | Wavelet networks are a new class of networks that combine the classic sigmoid neural networks (NNs) and the wavelet analysis (WA) | RUL prediction of bearing | [141] |
| **EMD+PSO+ SVM** | Hybrid of Physics-based model | Empirical Mode Decomposition performs operations that partition a series into 'modes' (IMFs; Intrinsic Mode Functions) without leaving the time domain. Particle Swarm Optimization is a computational method that optimizes a problem by iteratively improving a candidate solution concerning a given measure of quality. | Pump fault, RUL | [147] |
| **Fuzzy logic + Focused TLFN** | Knowledge-based + Data-Driven | In the Feedforward Neural network, the information moves in only one direction from the input nodes, through the hidden nodes (if any) and output nodes. Thus, there are no cycles or loops in the network. | Cutting tool monitoring, RUL | [148] |
| **Fuzzy Logic+ Kalman Filter** | Knowledge-based + Data-Driven | Kalman filtering is an algorithm that provides estimates of some unknown variables given the measurements observed over time | Actuator fault, RUL | [140] |
| **SVR+HMM** | Hybrid data-driven approach | The Support Vector Regression uses the same principles as SVM for classification. However, it allows us to define how much error is acceptable in our model and find an appropriate line to fit the data. | Bearing RUL | [43] |
| **MLP+RBF+KF** | Data-driven + physics-based approach | Multi-layer Perception is a class of Feedforward NN which refers to networks composed of multiple layers of perceptrons. A radial basis function is a real-valued function whose value depends only on the distance between the input and some fixed point, either the origin or another fixed point called the center. | RUL Prediction | [149] |
| **GPR+KF/PF** | Data-driven + physics-based approach | Gaussian process regression is nonparametric, Bayesian approach to regression that is making waves in machine learning. Particle filters or Sequential Monte Carlo (SMC) methods are a set of Monte Carlo algorithms used to solve filtering problems arising in signal processing and Bayesian statistical interference | RUL Prediction | [45] |
| **SK+ELM** | Data-driven + physics-based approach | The spectral kurtosis (SK) is a statistical tool that can indicate the presence of series of transients and their locations in the frequency domain | Bearing faults | [138] |

### 3.6.1. Machine Learning Styles

Machine learning emphasizes "learning," because of which there are different Machine learning techniques available. It is vital to know what type of Machine learning can be used in the fault diagnosis of rotating machines. These techniques are broadly classified as learning problems, hybrid-learning problems, statistical inference, learning techniques, and other important learning approaches [197, 198]. First are the three main learning problems in machine learning: supervised, unsupervised, and reinforcement learning [199]. The term "supervised learning" refers to an issue in which a model is used to learn a mapping between input instances and the target variable. In this case, class labels should be available in advance. Also, the class labels are extremely biased to normal class, whereas the anomalies appear rarely. Unsupervised learning refers to a set of issues in which a model describes or derives relationships from data. Here, the models treat instances that fit least to the majority as anomalies. Also, the model learns from partially labeled normal data and scale anomaly to the difference between an unseen pattern and the learned normal pattern. The difficulty here is that most of the machinery data is sequential and collected over a longer period. However, prediction over a longer sequence may be challenging as the length of the input sequence can vary. Reinforcement learning [200] is a type of online learning that simultaneously combines the learning and action phases and has a self-optimizing characteristic. The distinction between unsupervised and supervised learning is hazy, and several hybrid techniques include elements from both fields like semi-supervised, self-supervised, multi-instance learning, etc. Semi-supervised learning [201] employs unlabelled instances to aid in learning the probability distribution over the input space. It also simultaneously optimizes the prediction over labeled and unlabelled examples. Self-supervised learning [202] is a representation learning approach in which unlabelled data is used to construct a supervised task. Self-supervised learning is used to lower the expense of data labeling and use the unlabelled data pool. Contrastive learning [203], a type of self-supervised learning, identifies comparable and dissimilar objects for an ML model in a different way. The fundamental goal of contrastive learning is to develop representations that keep comparable samples near together while a large distance separates different samples. Contrastive learning is one of the most powerful techniques in self-supervised learning when working with unlabelled data. Multi-instance learning is a supervised learning problem in which individual instances are not labeled, but bags or groups of samples are. The process of arriving at a conclusion or judgment is known as inference. Both fitting a model and making a prediction are examples of inference in machine learning. It is the next category of learning, including techniques such as Inductive Learning, Deductive Inference, Transductive Learning, etc. Inductive learning [204] entails determining the result based on evidence. The deduction, also known as deductive inference [205], uses general rules to arrive at specific conclusions. In the realm of statistical learning theory, the term "transduction" or "transductive learning" [206] refers to predicting particular cases given specific examples from a domain. There are several strategies classified as learning techniques. It encompasses multi-tasking, active learning, online learning, transfer learning, and ensemble learning. Multi-task learning [207] is a form of supervised learning in which a model is fitted to a single dataset to solve numerous related tasks. Active learning [208] is an approach in which the model may ask a human user operator questions throughout the learning process to clarify ambiguity. Online learning entails using the data at hand and updating the model immediately before making a prediction or after the final observation. Transfer learning [209] is a form of learning in which a model is initially trained on one task and then utilized as the starting point for another activity. Ensemble learning [210] is a technique in which two or more models are fitted to the same data, and their predictions are pooled. Apart from these, there are many other techniques for machine learning such as meta-learning, continual learning, feature learning, federated learning, rule-based machine learning, multi-view learning, self-taught learning, deep learning, and many more. Learning algorithms that learn from other learning algorithms are called meta-learning [211] in machine learning. A model's capacity to learn continuously from a data stream is known as continual learning [212]. In reality, this means enabling a model to learn and adjust autonomously in production as new data arrives. Feature learning [213], also known as representation learning, is a collection of machine learning algorithms that allow a system to automatically find the representations needed for feature detection or classification from raw data. It eliminates the need for human feature engineering by allowing a machine to learn and use features to fulfill a given activity. Federated learning [214] is a technique for training machine learning algorithms across numerous edge devices without exchanging training data. Federated learning, as a result, provides a novel learning paradigm in which statistical techniques are taught at the network's edge. The rule-based machine learning approach of association rule learning identifies interesting relationships between variables in big databases. Multiview learning [195], also known as data fusion or data integration from various feature sets, incorporating several perspectives to increase generalization performance. Self-taught learning [215] is a new paradigm in machine learning. It uses labeled data belonging to the desired classes and unlabelled data from other, somehow similar classes. Many machine learning researchers have endorsed newer advances such as Deep Learning (DL) [163]. Due to its improved capacity to describe system complexity, DL has emerged as a viable computational tool for dynamic system prediction, overcoming the drawbacks of conventional techniques. In deep architecture, it is a machine learning approach that learns many levels of representations. Out of all the learning techniques, Supervised learning for classification (forecasts if the following n-steps have a chance of failure) and regression (to predict Remaining Useful Life), unsupervised learning (for anomaly detection), reinforcement learning, hybrid learning methods, transfer learning, ensemble learning, deep learning, meta-learning, ensemble learning,  etc. are widely used. Fig. 16 summarises all the machine learning techniques discussed above.

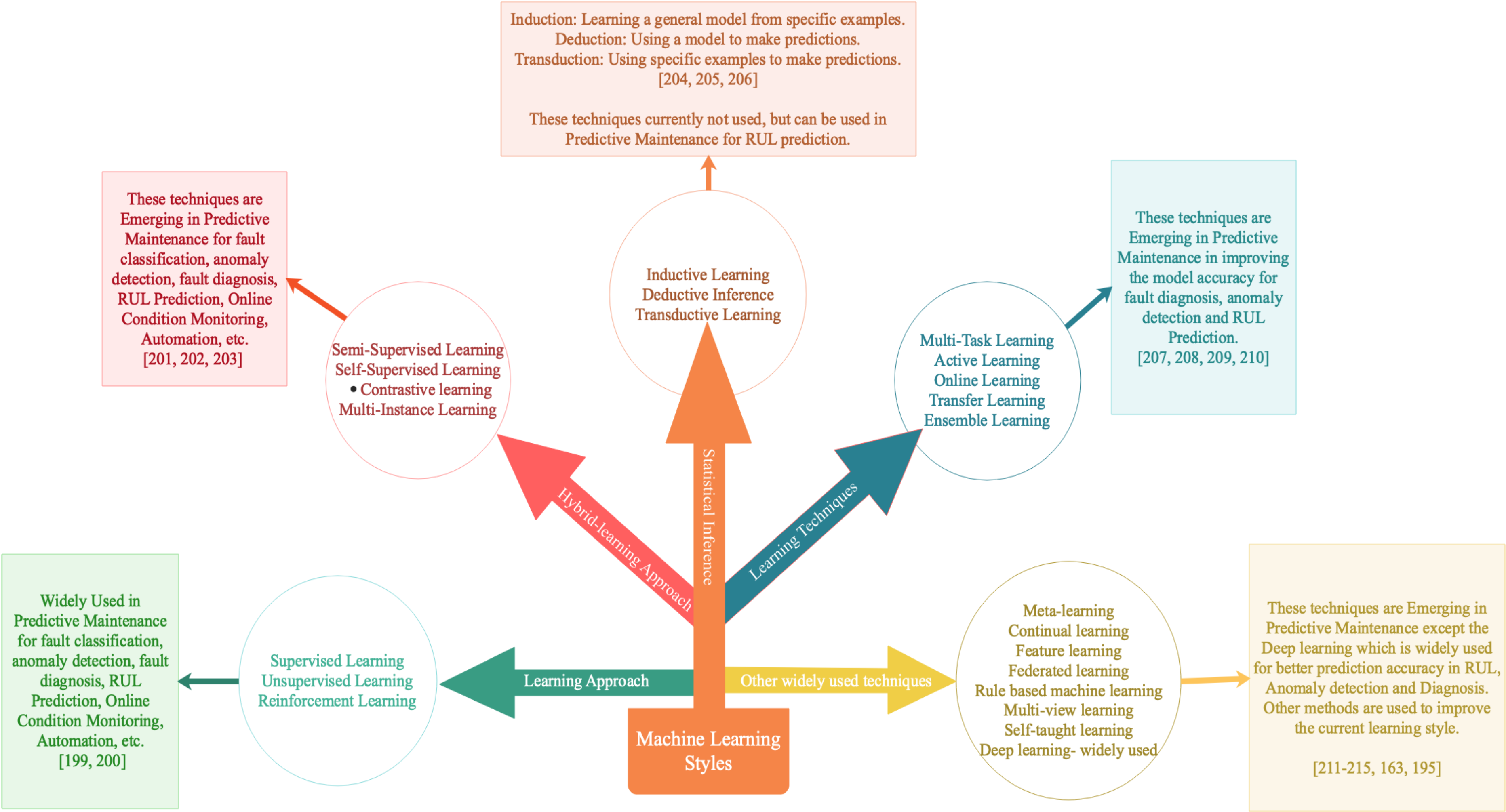

**Figure 16.** Overview of various Machine Learning Styles.

### 3.6.2. Steps to implement AI Model in fault diagnosis of rotating machines.

The implementation stack of AI for multiple fault diagnosis is depicted in fig.17. The figure is divided into several layers connected through a single directional road map. Each layer reveals a specific task accomplished via a particular set of either hardware or software or protocols or data stores. The first and the primary layer is the physical layer that includes the machine under test, the different sensors, the data communication medium, and the data acquisition system. Before moving ahead, it is essential to decide whether cloud computing or edge computing will be employed in fault diagnosis. Data pre-processing is the next important task that involves data storage, feature extraction, and data fusion. The next step is implementing the data analysis using ML/DL algorithms to achieve an accurate diagnosis. It includes training, testing, and validation of the algorithm and the results obtained. Once the diagnosis results are obtained, the final step is to convert the model to optimized C code to be implemented on the microcontroller. The advantages of using microcontrollers are: they consume low energy, are cheap, they are flexible, and most importantly, they have high security. Some widely used microcontrollers include Coral Dev Board, NVIDIA® Jetson Nano™ Developer Kit, Raspberry Pi 4 computer model B, etc.

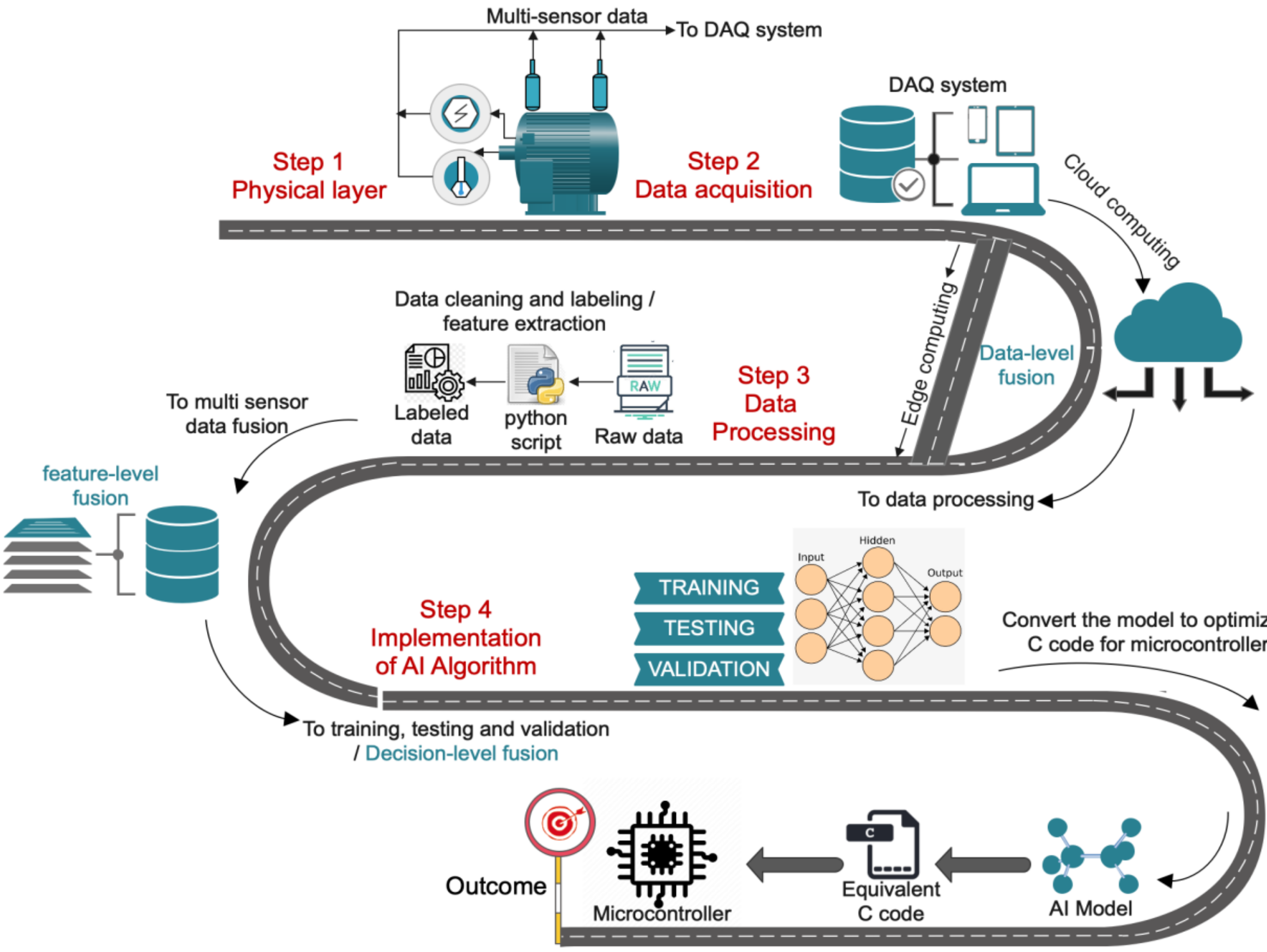

**Figure 17.** Implementation Stack of AI for Multi-Fault Diagnosis of Rotating Machines

## 4. DISCUSSION

With Artificial Intelligence (AI) advancement, a data-driven approach for predictive maintenance is taking a new flight towards smart manufacturing. This approach for multiple fault diagnosis in Industrial Rotating Machines is a never-ending research field. The literature review has developed new insight into the various aspects that a researcher should know before taking a step in this field. In this regard, let us discuss the survey outcome in the following subsection.

## 4.1. Survey Outcome

Let us analyze the survey outcome concerning the various aspects studied in the literature through Table 16. Finally, fig. 18 gives the summary of the literature review.

Table 16. Survey Outcome Discussion

| **Survey topic** | **Outcome** |
|---|---|
| Maintenance Strategy | Maximizing equipment reliability and facility performance while balancing the related resources used and, consequently, the cost is the goal of a good maintenance strategy. In this regard, there is a need to blend all the maintenance philosophies to provide the best strategy that would increase the RUL of the machine, Predictive maintenance being the core of all. |
| Data source | There are online datasets available. However, the best choice regarding validity and accuracy is to yield an accurate diagnosis, design a test setup, and collect data manually to train the AI models. |
| Sensors | Sensor fusion refers to the capacity to combine data from various sources and sensors to create a single model or representation of the world surrounding the machinery. Because the strengths of the many sensors are balanced, the final model is more accurate. |
| Data Acquisition | There are a variety of data acquisition options available. One must choose a suitable system based on the application and compatibility. It is always better to decide the data acquisition method before the purchase of sensors and other apparatus. |
| Signal Processing | Feature selection is critical in signal processing. For non-stationary signals, which are very common while machinery failure, features from the time-frequency domain need to be used for better results. |
| AI techniques | Hybrid data-driven approaches boost the advantages and conceal the disadvantages of different independent AI techniques to yield a better diagnosis. |

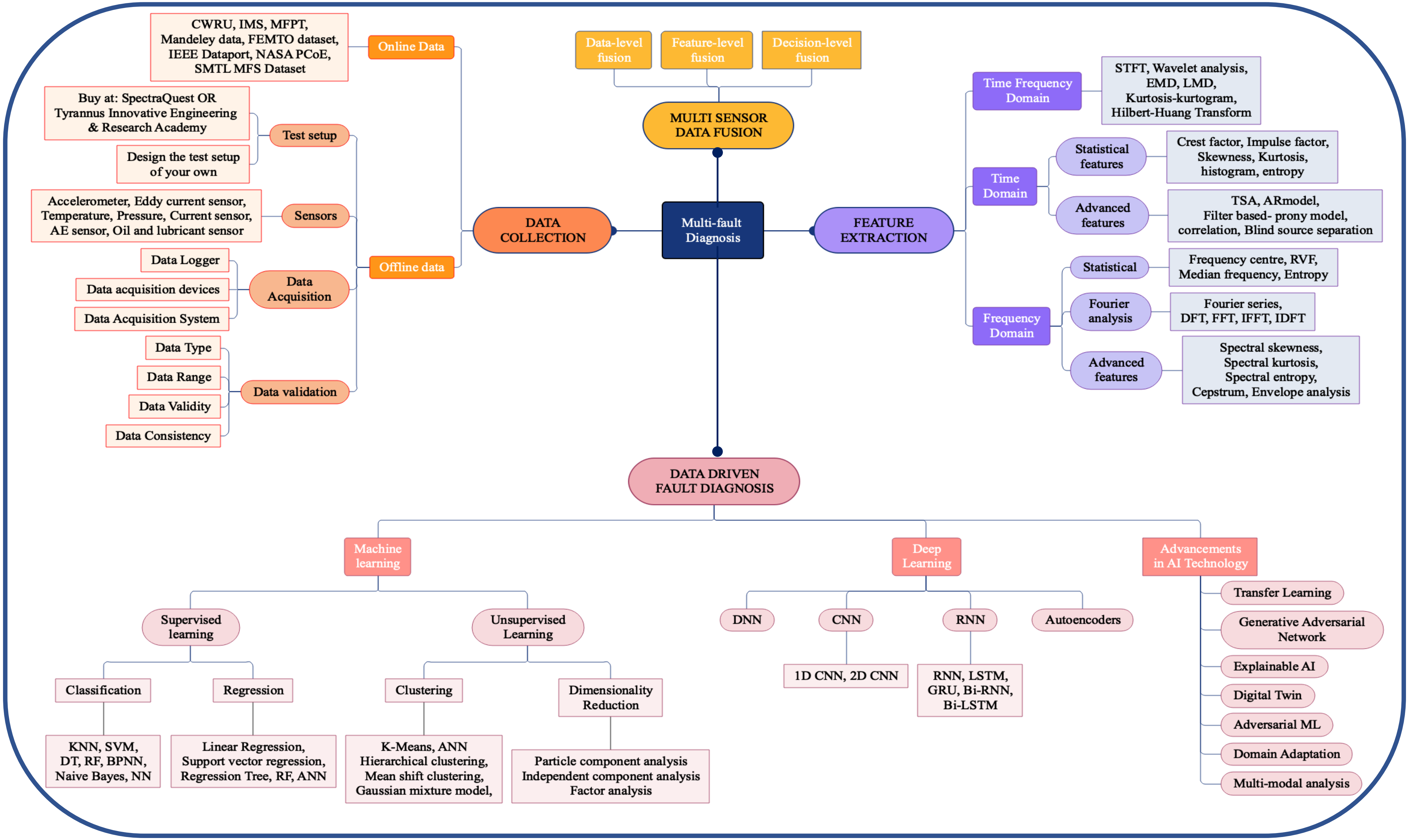

**Figure 18.** Complete Summary of Literature Revie

## 4.2. Challenges and limitations in multi-fault diagnosis

The challenges and limitations of multi-fault diagnosis are discussed in this paper that may aid in further research. Table 17 gives a systematic overview of the challenges and limitations.

Table 17. Challenges and Limitations of Multi-Fault Diagnosis in Industrial Rotating Machines

| Challenges | Issues | Impact on Multi-fault diagnosis |
|---|---|---|
| **Data related challenges** | Data Availability | Machines with a wide range of dynamic operating circumstances might lack complete data with a comparable distribution for fault diagnosis. |
| | Data Accessibility | Failure to access high-quality data in a defined format for AI models may harm the models' performance. |
| | Data quality | Due to a lack of access to high-quality data, firms and governments may be forced to develop subpar AI systems for fault diagnosis systems. |
| | Data acquisition/ transmission | If appropriate and high-speed communication and data acquisition systems are not available, critical information may be truncated. |
| | Single sensor data | If only one type of data is used to train the AI model, it may not be effective. The diagnosis is incomplete by depending on the single type of data. |
| | Cost for data collection | Due to the high cost of machinery parts, it is challenging to run the machinery until failure for data collection. |
| **AI Model related Challenges** | Class Imbalance | Due to the failure of industrial sensors, data is missing or partial. |
| | Model Interpretability | Inability to understand why a particular diagnosis was given at a specific point in time due to a lack of model interpretability |
| | adversarial perturbation | In the case of essential equipment, AI models are vulnerable to adversarial perturbations, which might further weaken the credibility of fault diagnosis |
| | Inefficiency of the model for similar (not exact) data. | The data is never the same for training and testing due to broad operating conditions and circumstances in an Industrial environment. |
| **Network and security** | Cloud computing | Concerns about security, privacy, and real-time performance may arise while using cloud computing. As a result, it is necessary to verify the dispersed resource sharing of critical industrial data. |
| | Privacy of the data | Designing safe and hack-free durable intelligent diagnosis is a significant problem due to its complexity. |

## 4.3. Advancements in Data-driven Multi-fault diagnosis in the context of Industry 4.0

Artificial intelligence progress is genuine and occurring with the technology finding extensive use in nearly every industry. In such industries, Predictive maintenance, which is a method of gathering, analyzing, and utilizing data from different industrial sources such as machines, sensors, and switches, uses intelligent algorithms to analyze data to predict equipment failure before it occurs. Companies are already using continuous monitoring technologies like the Internet of Things (IoT). However, the key to increasing company efficiency is going beyond simply monitoring the output of multiple technologies and using powerful techniques in machine learning to act on real-time insights and improve upon significant issues faced by current techniques. Predictive maintenance is at the heart of industrial innovation, and it entails rethinking and optimizing the whole maintenance approach from top to bottom. AI-led approaches have addressed some open issues in the multi-fault diagnosis of rotating machines. These approaches include generative adversarial networks, explainable AI, transfer learning, domain adaption, digital twin, adversarial machine learning, and domain adaptability. Figure 19 depicts some of these challenges and the solutions given by these approaches.

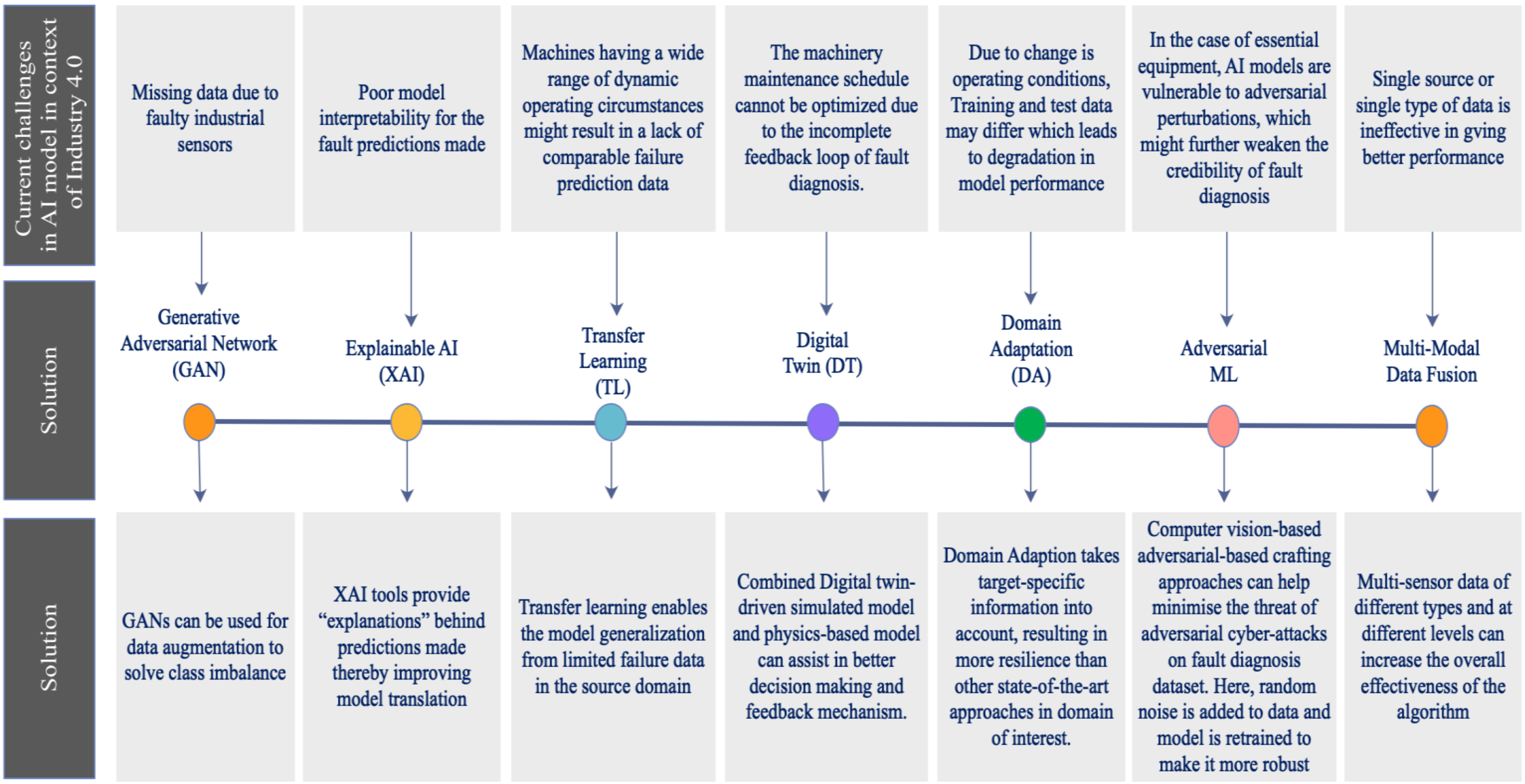

**Figure 19.** Summary of Challenges and the corresponding Advancements in AI Models

### 4.3.1. Generative Adversarial Networks (GAN)

Sensors installed to collect machine condition data might malfunction due to inadequate power supply. The data might differ due to constantly changing conditions in the manufacturing industries, leading to data shortage. The highly unbalanced training data, the extremely high cost of obtaining more failure examples, and the intricacy of the failure patterns can also add to further challenges. GAN starts by using two GAN networks to produce training samples and develop an inference network that may be used to forecast new sample failures. GAN is used to synthesize virtual artificial data, an unsupervised learning approach involving the interaction of two neural networks [174]. The GAN is made up of two parts: a generator and a discriminator. The generator's job is to create synthetic data as close to the actual sample as feasible, while the discriminator's job is to separate real samples from fake samples as much as possible. The schematic of GAN is as shown in fig. 20. [174] Mode collapse, Non-conversions, instability, and Evaluation matrix issues are some of the challenges in implementing GANs [175]. One possible study direction is to investigate how these approaches might be applied to more complicated datasets containing vibration and time-series data [190].

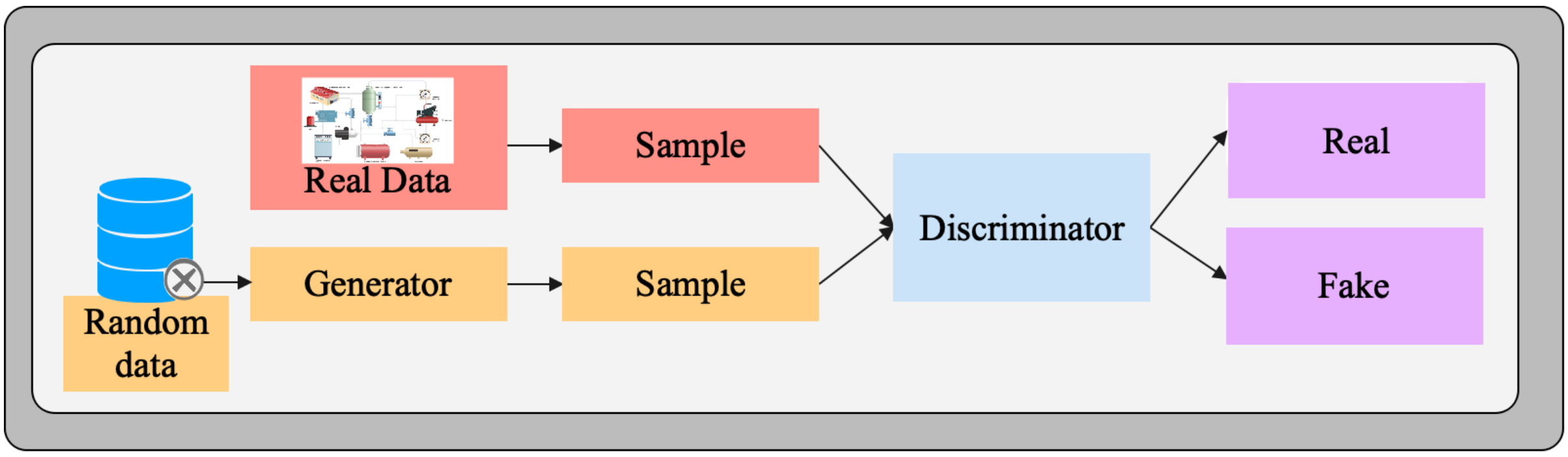

**Figure 20.** GAN Architecture.

### 4.3.2. Explainable AI (XAI)

As the industries are turning towards automation, users are progressively outsourcing more jobs to computers. Users may find it challenging to grasp such complicated systems since they are generally developed utilizing "black box" Artificial Intelligence (AI). Traditional black-box AI algorithms might be more transparent by offering explanations of the findings or building explainable and interpretable AI solutions for Industry 4.0 applications. Future research into Explainable AI (XAI) in predictive maintenance is critical since interpretability and clarity are essential aspects for building credibility and security. Galanti et al. [176] suggested an explainable AI solution for process monitoring. On real industrial benchmark datasets, the suggested solution is assessed. Rehse et al. [177] have also examined the challenges and possible applications of explainable AI in Industry 4.0 in great depth. Defining the relevant evaluation mechanisms, procedures, measures, and methods are some of the challenges in Explainable AI. The quality, usefulness, and satisfaction of the explanations, and the influence of explanations on the model's success are all factors to consider when evaluating the XAI [191]. Fig. 21 [178] shows the architecture to implement XAI.

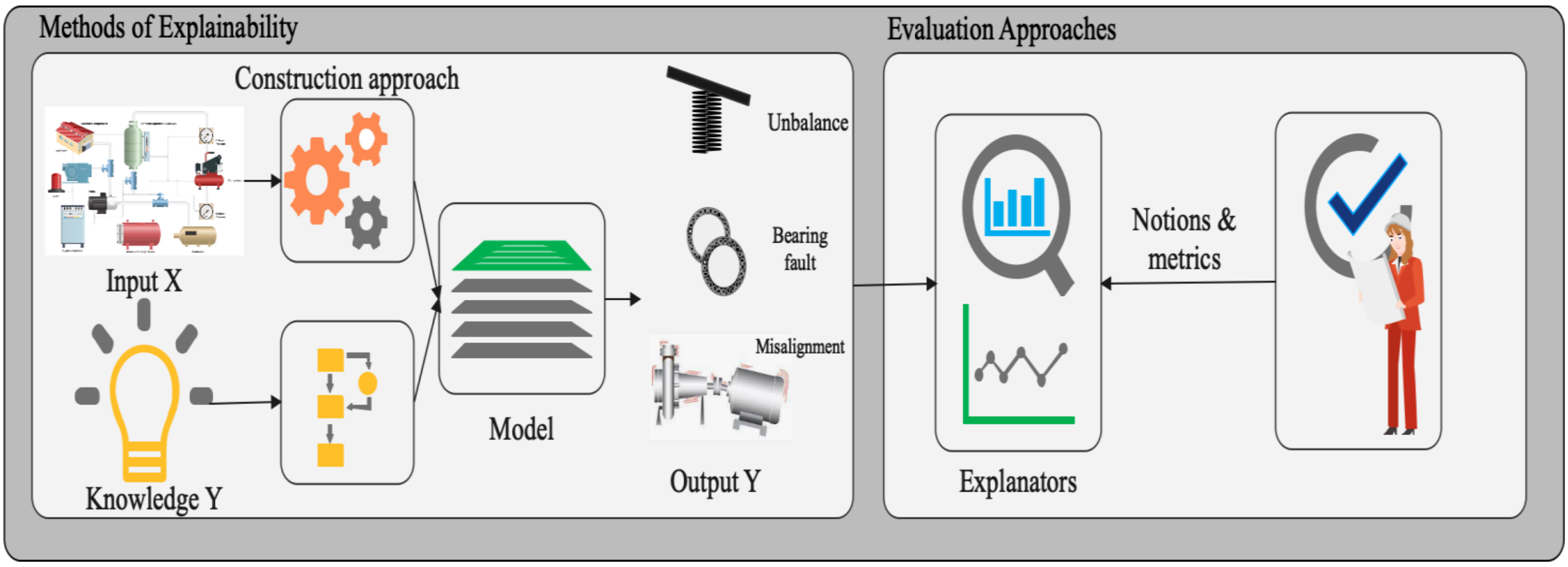

**Figure 21.** XAI Architecture.

### 4.3.3. Transfer Learning (TL)

Transfer learning is the concept of breaking out from the isolated learning paradigm and using what has been learned to tackle related problems. Due to constantly changing conditions in manufacturing units, the data is never the same, but it can be similar. Transfer Learning (TL) methods can enhance model accuracy for pre-and post-model deployment in dissimilar data distribution across the source and target domains. Fig 22 [179] is an architecture of Digital Twin assisted transfer Learning. The paper's authors devised a technique known as Digital-twin-assisted Fault Diagnosis using Deep transfer learning (DFDD). In the first phase, the potential problems that are not considered at design time were discovered by front running the ultra-high-fidelity model in the virtual space. In contrast, a Deep Neural Network (DNN) based diagnosis model was fully trained. In the second phase, the previously trained diagnosis model was migrated from the virtual to physical space using Deep Transfer Learning (DTL) for real-time monitoring and predictive maintenance. Implementing transfer learning for multiple fault diagnosis is currently an emerging research area.

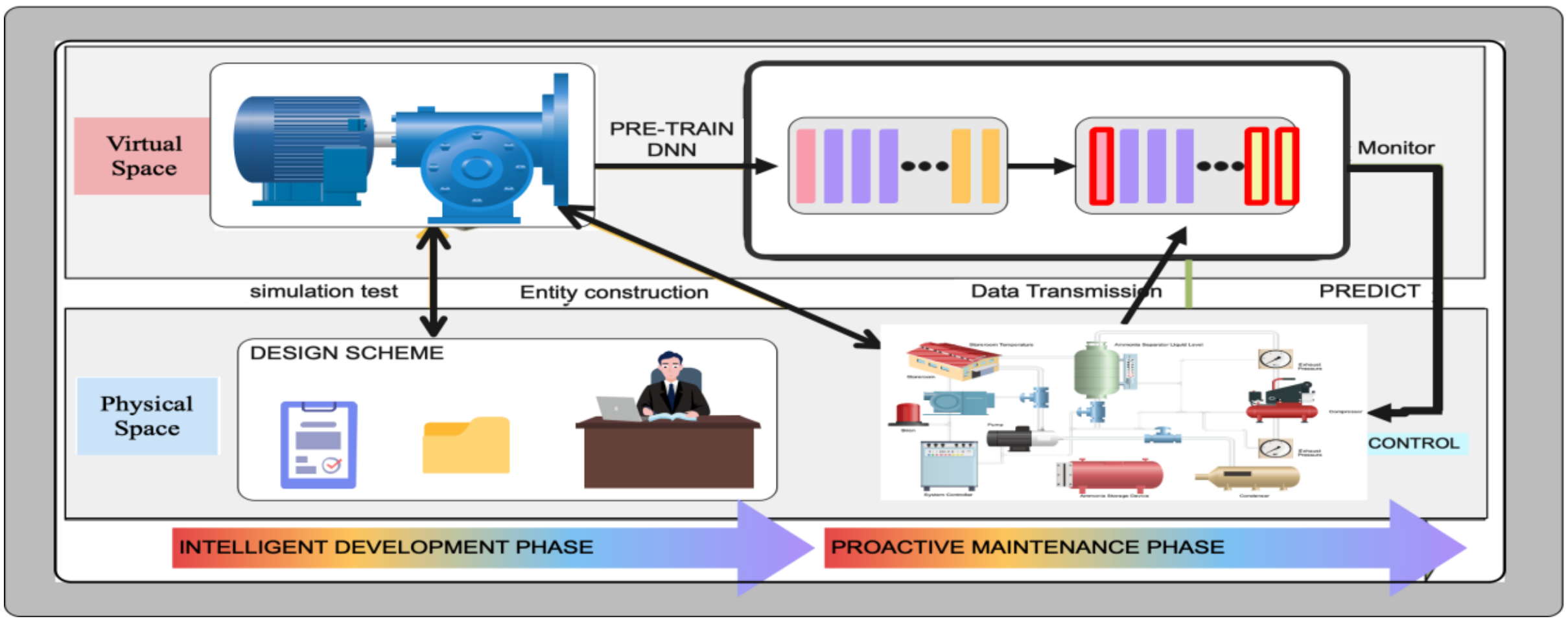

**Figure 22.** Digital Twin Assisted Transfer Learning.

### 4.3.4. Digital Twin (DT)

The goal of a Digital Twin is to construct a very accurate virtual replica of a physical system or process to mimic system behavior, condition monitoring, unusual pattern identification, system performance reflection, and future trend projection [9]. The ability to produce simulated data is one of Digital Twin's features. Infinite cycles of situations must be run in a simulated environment. The simulated data collected may subsequently be utilized to successfully train a naive AI model on the various elements of manufacturing operations. The digital twin's next skill is finding, planning, and testing new features that may be utilized to supplement data operations in a machine learning process. Finally, a digital twin can help with data augmentation for imbalanced datasets. DT is extensively used for predictive maintenance, fault diagnosis, detecting system abnormalities, inferring product quality, real-time system monitoring, and so on [179-181]. Fig 22 [179] is an architecture of Digital Twin assisted transfer Learning. However, digital twin challenges are accurately capturing physical properties, project collaboration, automatic real-time updating, conflict detection and resolution, and interaction with digital and physical items. Digital Twin enabled fault diagnosis can be implemented, which can help leverage the real-time and wholistic information related to machinery giving feedback between the real and digital world at every stage.

### 4.3.5. Domain Adaptation (DA)

Fault diagnostic models are created with a specific machinery configuration in mind, but there may be a situation when they need to be applied to a different machinery design. The new equipment configuration may affect the model prediction accuracy, which is typically different from the prior one. The change in data can also happen due to collecting the training and test sets from different sources or having an outdated training set due to data change over time. In this case, there would be a discrepancy across domain distributions, and naively applying the trained model on the new dataset may cause degradation in the performance. Domain Adaption (DA) can aid in such issues and implement efficient extraction of features from unlabelled equipment data, which is a prevalent problem in most real-time industrial applications [184]. The author in [183] presented a unique RUL prediction technique based on a deep convolutional neural network (DCNN) coupled with Bayesian optimization and adaptive batch normalization (AdaBN). The results also demonstrate that the prediction model's domain adaptability capacity has improved. One of the most common techniques to transfer learning is domain adaptation, which assumes that the label spaces of the source and destination domains are similar. The architecture and concept of domain adaptation presented by [185] are shown in fig.23 [185]. Heterogeneous unsupervised domain adaptation is a naïve emerging field to be worked on in the future.

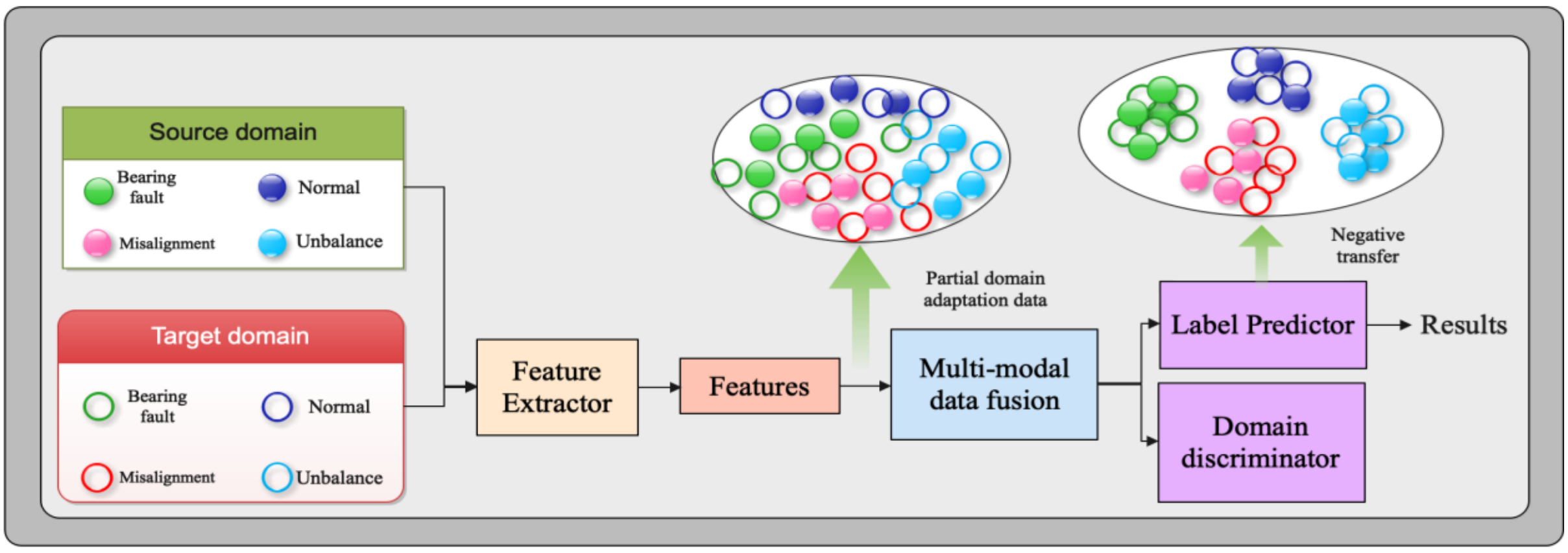

**Figure 23.** Adversarial Domain Adaptation Architecture.

### 4.3.6. Adversarial ML

While certain machine learning models are good at predicting predictions, they may not be good at detecting unlawful intrusions. Adversarial Machine Learning (AML) models protect the model structure from adversarial assaults that endanger the predictive maintenance framework's resilience. [182] provides a list of adversarial ML attacks, from which the top four attacks with the most industrial impact are poisoning attack, Evasion attack, Trojan attack, and model extractions/ stealing attack. The findings in [186] show that CBM systems are vulnerable to adversarial machine learning assaults, necessitating security measures. According to the timing of the attack, the approaches employed by hackers for adversarial machine learning may be classified into two categories: Data Poisoning: To deceive the output model, the attacker modifies the labels of certain training input instances. Model Poisoning: After the model is constructed, the hacker causes it to provide incorrect labeling by utilizing a perturbed instance. The adversarial machine learning attack is depicted in Figure 24 [196, 187]. Blockchain-enabled [219] reliable machine condition monitoring and fault diagnosis model can be a domain for future research.

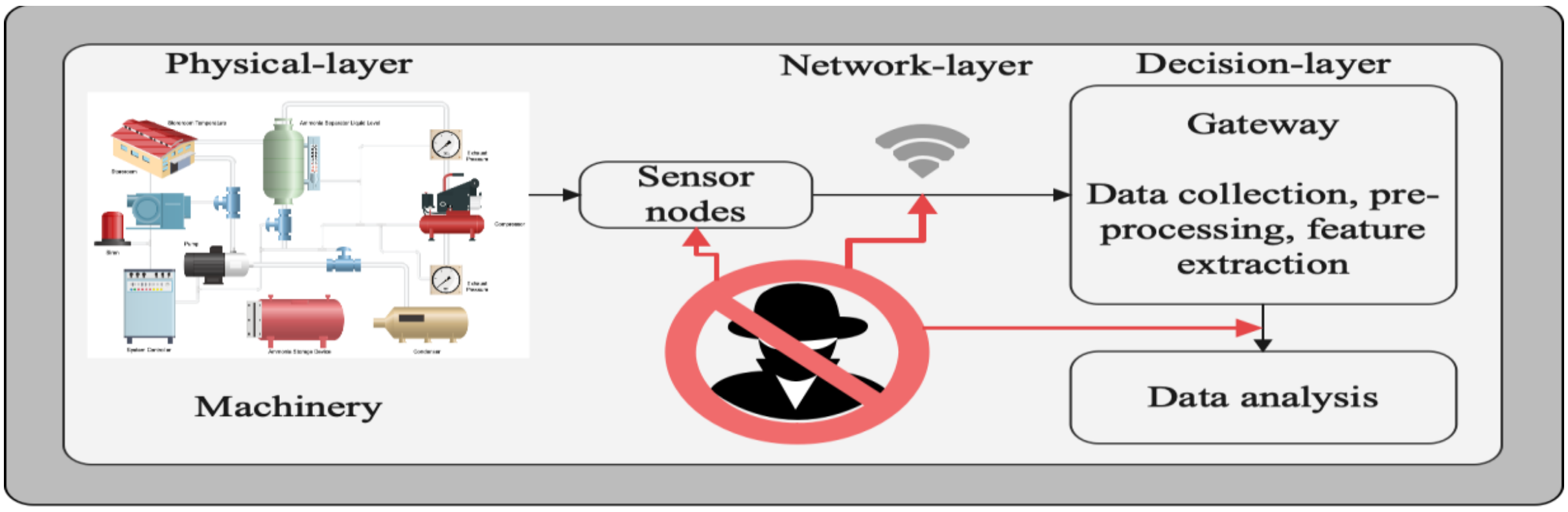

**Figure 24.** Example of Adversarial ML Attack.

### 4.3.7. Multi-Modal Data Fusion

Compared to single modalities (i.e., unimodal) systems, multimodal deep learning systems employ several modalities, including vibration, temperature, pressure, AE, etc., and perform better. Representation, translation, alignment, fusion, and co-learning are all aspects of multimodal machine learning. Some of the challenges in the multimodal analysis include missing modalities, noisy data, a lack of annotated data, inaccurate labeling and scarcity during training-testing, domain adaptability for diverse datasets, and interpretability of results [188]. Figure 25 [189] shows the multi-sensor data collection for Multi-Modal Data Fusion (MMDF). From a future perspective in the context of Industry 4.0, the majority of multi-modal data in smart manufacturing setups must be gathered in dynamic settings, implying that the data itself varies. Also, it should rate the essential modalities in the diagnosis of the fault.

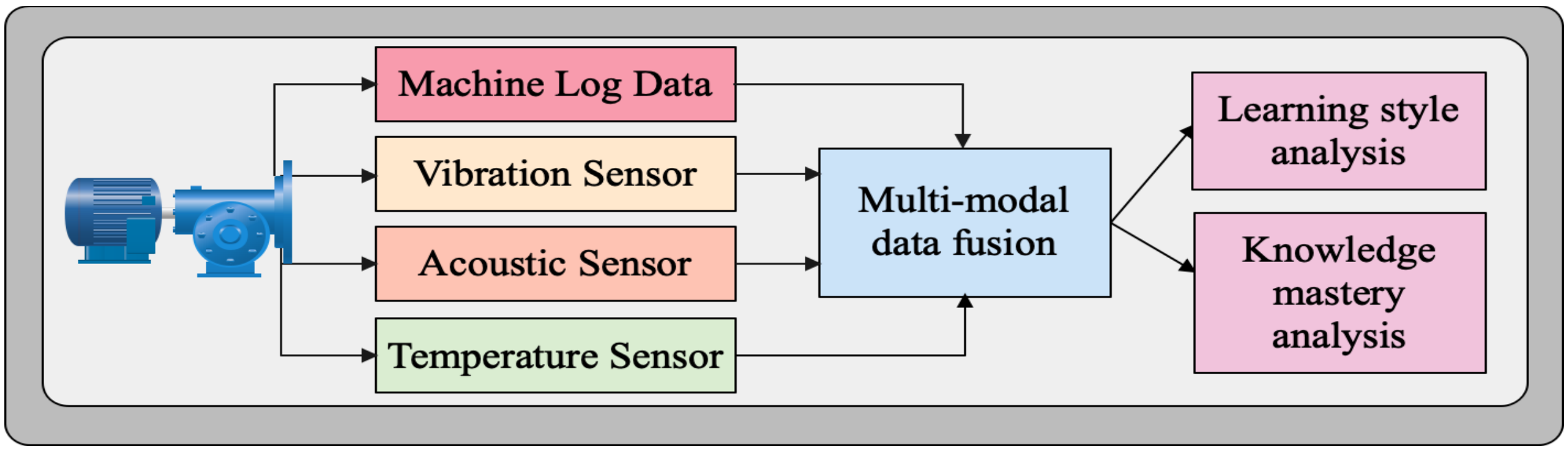

**Figure 25.** The architecture of Multi-modal Perception with Data Fusion and Analysis.

## 5. RECOMMENDATIONS FOR FUTURE WORK

Apart from the above-mentioned future research work in each current advancement, the authors would like to suggest a few other possible research directions in Data-Driven Multi-fault diagnosis of Industrial rotating machines. Table 18 describes the research gap identified and the corresponding future direction for the upcoming research. It should be mentioned that the range of this study is limited to a sample of research journals and keywords. It is proposed that the procedure outlined here be repeated for the same research theme in the future, with more databases taken into account.

Table 18. Research Gap and Future Direction for Multi-Fault Diagnosis in Industrial Rotating Machines.

| Sr. No. | Research Gap | Future Direction |
|---|---|---|
| 1 | Robust data collection is lacking as the researchers mainly focus on either online data (which is unimodal) or not considering the multi-modal aspect of the data. | There is a need to use multi-sensor data fusion or multimodal analysis to experiment with multi-fault diagnosis in predictive maintenance. |
| 2 | There is extensive research on bearing faults. However, industrial rotating machines also have faults like unbalance, misalignment, looseness, etc., which are related to each other. They are equally essential to be considered along with bearing faults for the study to be complete. | Research should now focus on the multi-fault aspect of fault diagnosis in rotating machines. |
| 3 | Exhaustive research has been implemented using standard datasets available online. However, real-time industrial conditions are continuously changing. That is why most AI models fail when they are implemented in a real-time industrial environment. | There is a high need to use domain Adaptation and transfer learning to the models to reduce the domain dependence of the model. Also, one must consider industrial conditions while working on their test setup (avoid online data) generating benchmark datasets. |
| 4 | Validating the proposed algorithm or model in the real-time industrial environment lacks most of the articles published. | There is a need to prove the algorithm's accuracy by validating it in a real-time industrial environment. |
| 5 | Research in Digital twin + predictive maintenance is very little and still in the phase of development. | Digital twin, emerging technology and very effective in predictive maintenance, is a new challenge for upcoming researchers. |
| 6 | Many researchers use a data-driven model or a model-based technique to compute the fault diagnosis, which may contain prediction mistakes due to individual models' uncertainty. | A hybrid data-driven strategy combined with hybrid decision-making algorithms might reduce fault prediction mistakes. |
| 7 | Changes in sensor operating conditions, disruption due to big equipment starting, high-frequency interference, and other factors taint sensor signals. As a result, it is difficult to eliminate or filter noise from raw signals. Also, the dataset is imbalanced or missing. | Integrated de-noising based on energy-correlation analysis and wavelet transform packet can be used to overcome industrial sensor signal de-noising. GANs can also be implemented to generate missing data or imbalanced data synthetically. |
| 8 | Model optimization is needed. | Reinforcement learning, an ML technique where the model learns through trial and error to choose the best course of action, can be used for model optimization. |
| 9 | As discussed earlier, cloud computing is prone to several security issues where AI/ML models can be attacked in several ways. | The development of adversarially resilient AI/ML models for industrial applications is still a topic of research that has to be addressed. Balance of cloud, as well as Edge computing, can also be one of the solutions. Edge |
| | | computing allows data to be processed locally (i.e., near the collecting devices), reducing bandwidth and latency significantly, and it is also more secure. |
| **10** | There is also a need to know why an AI model has come up with a particular decision. | Explainable AI is also a potential future direction to focus on, as discussed earlier. |
| **11** | IoT devices are anticipated to exceed several trillion in the following years, posing significant performance and data monitoring problems [192]. | Another critical challenge that might be investigated in the future is scalability. |
| **12** | Energy and hardware constraints are two of the most critical roadblocks to ML adoption in Industry 4.0. | Research is required to improve and optimize energy usage and conservation in IIoT devices [192]. In addition, real-time ML through online or incremental learning can also be further studied. |
| **13** | Representation of a large amount of data generated from multiple sensors is challenging to manage. | Knowledge Graphs (set of datapoints linked by relations) is a powerful way of representing data that can be built automatically and can then be explored to reveal new insights about a domain [193]. |

## 6. CONCLUSION

This study focuses on data-driven predictive maintenance in industrial rotating machinery for multi-fault detection. According to the available literature, multi-fault diagnosis is still a developing field with much room for advancement in Industry 4.0. The study examines several open research topics that researchers in this field are grappling with by implementing a systematic literature review on a Data-driven approach for multi-fault diagnosis of Industrial Rotating Machines. This review was implemented using the "Preferred Reporting Items for Systematic Reviews and Meta-Analysis" (PRISMA) method. The article discusses all aspects of the physical layer: the machinery and the sensors, followed by the various data acquisition methods employed to collect the data. There is also a systematic analysis of available online datasets in this field. The paper has also focussed on signal processing techniques, mainly feature extraction methods such as time domain, frequency domain, and time-frequency domain features. Authors have also covered the topic of multi-sensor data fusion at different levels, viz., the data-level, feature-level, and decision-level fusion. There is also a great focus on various data-driven and hybrid approaches implemented by the previous researchers. Covering the recent advancements in a data-driven approach for multi-fault diagnosis, the authors have identified the significant challenges faced in this domain and given the corresponding solution. Finally, the authors conclude the paper with the infographic diagram covering all the above aspects and addressing the future scope in the domain. The authors have tried to evaluate all aspects of data-driven multi-fault diagnosis in predictive maintenance and believe that this survey will assist the researchers, the essential background in future research directions.

## Declarations

Funding: NA
Conflicts of Interest: NA
Availability of data and material: NA Code availability: NA
Authors' contributions: NA
Ethics approval: NA
Consent to participate: NA
Consent for publication: NA